\documentclass[journal, 10pt]{IEEEtran}
\IEEEoverridecommandlockouts

\usepackage{etoolbox}
\usepackage{capt-of}
\makeatletter
\let\NAT@parse\undefined
\apptocmd\@maketitle{{\eyecatcher{}\par}}{}{}
\makeatother
\usepackage{amsmath,mathtools}
\usepackage{amsfonts}
\usepackage{amssymb}
\usepackage{balance}
\usepackage{cite}
\usepackage{graphicx}
\usepackage[skip=2pt,font=small]{caption}
\usepackage{subcaption}
\usepackage[numbers, square, comma,compress]{natbib}
\usepackage{siunitx}
\usepackage{color}
\usepackage[table]{xcolor}
\usepackage{booktabs}
\usepackage{footmisc}
\usepackage{dblfloatfix}
\usepackage{url}
\usepackage{nth}
\usepackage{animate}
\usepackage{comment}
\usepackage{hyperref}
\usepackage{pgfplots}
\usepackage{tikz}
\usetikzlibrary{arrows}
\usepackage[normalem]{ulem}
\usepackage{paralist}
\usepackage[ruled,vlined]{algorithm2e}
\usepackage{booktabs}
\usepackage{siunitx}
\usepackage[utf8]{inputenc}
\usepackage{pgfplots}
\usepackage{multirow}
\usepackage{amsthm}
\usepackage{enumitem}
\usepackage{graphicx}
\usepackage{subcaption}
\usepackage{multicol}
\usepackage[switch,columnwise]{lineno}

\DeclareUnicodeCharacter{2212}{−}
\usepgfplotslibrary{groupplots,dateplot}
\usetikzlibrary{patterns,shapes.arrows}
\pgfplotsset{compat=newest}

\newcommand{\rom}[1]{(\expandafter{\romannumeral #1\relax})}

\definecolor{royalazure}{rgb}{0.0, 0.22, 0.66}
\definecolor{mayablue}{rgb}{0.45, 0.76, 0.98}

\definecolor{somegray}{rgb}{0.5, 0.5, 0.5}
\newcommand{\darkgrayed}[1]{\textcolor{somegray}{#1}}

\makeatletter 

\newcommand*\titleheader[1]{\gdef\@titleheader{#1}}
\AtBeginDocument{%
  \let\st@red@title\@title
  \def\@title{%
    \vskip-2.0em
    \bgroup\normalfont\large\centering\@titleheader\par\egroup
    \vskip0.59em\st@red@title}
}
\makeatother
\titleheader{\darkgrayed{This paper has been accepted for publication at Nature Scientific Reports, 13, 9727 (2023). \\
\protect\url{https://doi.org/10.1038/s41598-023-36775-0}}}

\newcommand\eyecatcher{
\centering
\vspace{8pt}
\captionsetup{type=figure}\setcounter{figure}{0}
\includegraphics[width=0.75\linewidth]{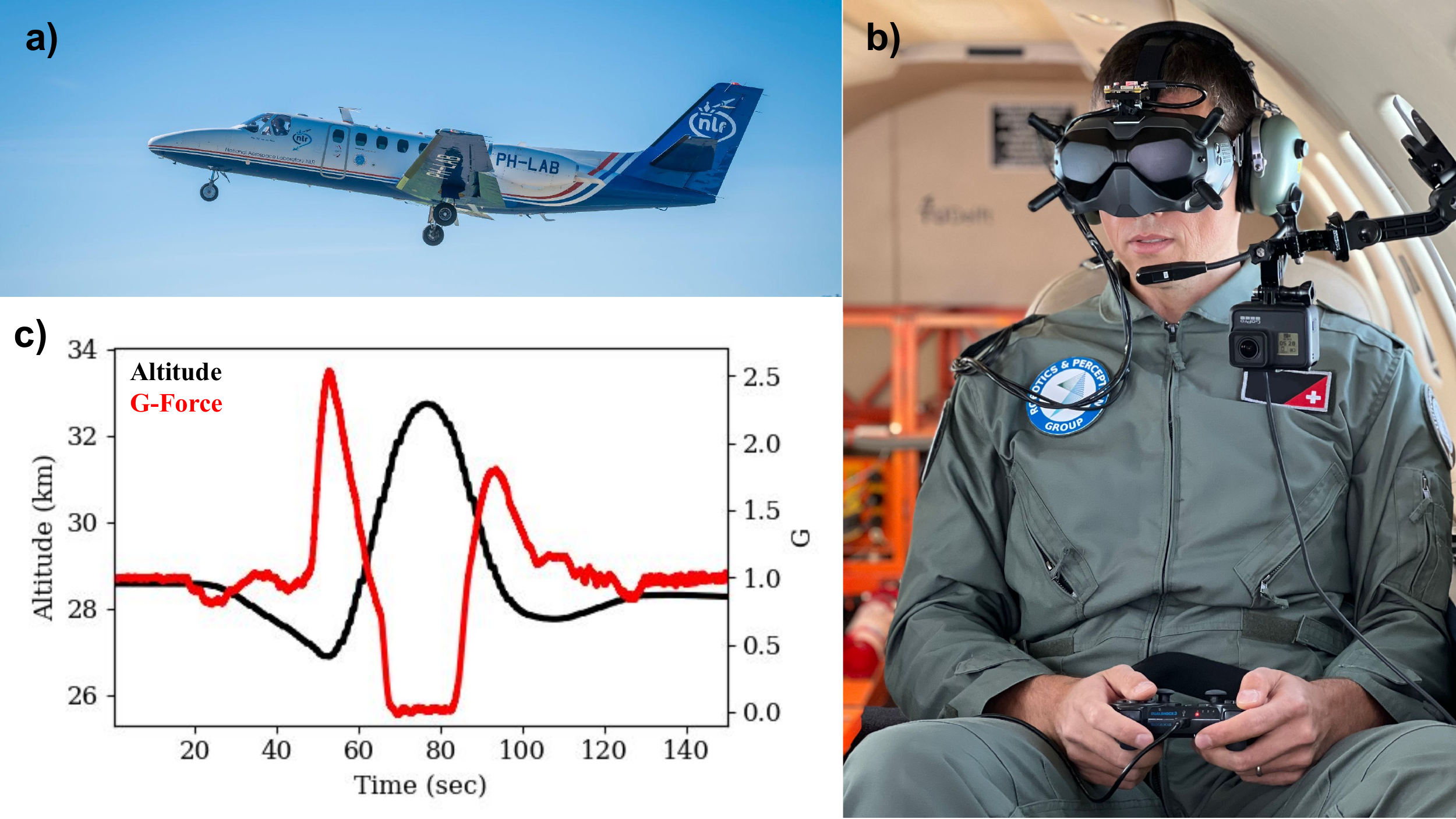}
\captionof{figure}{a) Cessna Citation II airplane used for parabolic flight. b) Research subject aboard the plane. c) Plane altitude (black) and resulting g-forces (red) as a function of time during an exemplar parabola maneuver.}
\label{fig:eyecatcher}
\vspace{-20pt}
}

\title{Microgravity induces overconfidence in perceptual decision-making}

\begin{document}


\author{Leyla Loued-Khenissi*, Christian Pfeiffer*, Rupal Saxena, Shivam Adarsh, Davide Scaramuzza
        \thanks{* Equal contribution. LLK is affiliated with the Laboratory for Behavioral Neurology and Imaging of Cognition, Neuroscience Department, Medical School, and the Neuro-X Institute, École Polytechnique Fédérale de Lausanne, Geneva, Switzerland. CP, RS, SA and DS are with the Robotics and Perception Group, University of Zurich, Switzerland (\protect\url{http://rpg.ifi.uzh.ch}).
        This work was supported by the Foundation for Research in Science and the Humanities at the University of Zurich (Grant number: STWF-22-004), Swiss Sky Lab Foundation, the Swiss National Science Foundation (SNSF) through the National Centre of Competence in Research (NCCR) Robotics and the European Research Council (ERC) under grant agreement No. 864042 (AGILEFLIGHT).}
}

\maketitle

\begin{abstract}

Does gravity affect decision-making? This question comes into sharp focus as plans for interplanetary human space missions solidify. In the framework of Bayesian brain theories, gravity encapsulates a strong prior, anchoring agents to a reference frame via the vestibular system, informing their decisions and possibly their integration of uncertainty. What happens when such a strong prior is altered? We address this question using a self-motion estimation task in a space analog environment under conditions of altered gravity. Two participants were cast as remote drone operators orbiting Mars in a virtual reality environment on board a parabolic flight, where both hyper- and microgravity conditions were induced. From a first-person perspective, participants viewed a drone exiting a cave and had to first predict a collision and then provide a confidence estimate of their response. We evoked uncertainty in the task by manipulating the motion's trajectory angle. Post-decision subjective confidence reports were negatively predicted by stimulus uncertainty, as expected. Uncertainty alone did not impact overt behavioral responses (performance, choice) differentially across gravity conditions. However microgravity predicted higher subjective confidence, especially in interaction with stimulus uncertainty. These results suggest that variables relating to uncertainty affect decision-making distinctly in microgravity, highlighting the possible need for automatized, compensatory mechanisms when considering human factors in space research.

\end{abstract}
\noindent
\section*{Dataset} The dataset can be downloaded at \protect\url{https://osf.io/zqfw4/}.

\section{Introduction}

The most salient interplanetary target of human space exploration is Mars, where unmanned vehicles, including NASA’s Ingenuity helicopter have already been sent \cite{schroeder_nasas_2020}, \cite{witze_lift_2021}. Human space flight missions must incorporate concerns on human factors associated with space exploration \cite{oluwafemi_review_2021}, notably in the effects of varying gravity conditions on human behavior \cite{moore_long-duration_2019}, \cite{pagnini2023human}, including decision-making. An unavoidable feature of decision-making is uncertainty, which is integrated computationally in human behavior brains and behavior \cite{loued-khenissi_information_2020}, \cite{loued2022effect} by way of internally held priors. Gravity is a strong sensory prior \cite{merfeld_humans_1999}, \cite{jorges_gravity_2017} that can affect perception and cognition \cite{grabherr_effects_2010}, \cite{Pfeiffer2014TheVS}. It is crucial then to understand if varying gravity conditions affect a human’s ability to integrate uncertainty in decision-making. By formalizing any such differences, through experimentally-induced alterations of gravity (Fig. \ref{fig:eyecatcher}), we can anticipate a potentially labile feature of human performance in space.

\subsubsection{Uncertainty in human brains and behavior}
Uncertainty is a ubiquitous feature of the environment. Studies in cognitive neuroscience have found that both neural signals and behavior are well fit by various formal models that integrate uncertainty \cite{spratling_review_2017}, including mean-variance \cite{loued-khenissi_anterior_2020} and Bayesian frameworks \cite{weilnhammer_neural_2018}. These algorithms can be embedded into a predictive coding model \cite{rao_predictive_1999}, a computational account of general brain function. Within this framework, uncertainty is resolved via (statistical) inference. Subjective post-decision reports can aid in assessing objective uncertainty-related variables. Agents overtly gauge their performance via metacognition, the process of reflecting on decisions, that gives rise to a post-decision uncertainty dubbed confidence \cite{fleming_metacognition_2012}. In healthy humans, subjective confidence is tied to objective, prediction uncertainty \cite{pereira_evidence_2021}, \cite{atiya_explaining_2021}. Since gravity act as a strong, perhaps intrinsic prior \cite{barnett-cowan_gravity-dependent_2018}, its absence may alter cognition and metacognition \cite{arshad_cognition_2022},  \cite{ferre_vestibular_2020}, \cite{Pfeiffer2014TheVS}. 

\subsubsection{Human perception, vestibular system and gravity}
Humans perceive and act based on a process of a cross-modal integration of information \cite{deneve_bayesian_2004}, \cite{noel_cognitive_2022}. For instance, both body position and visual cues are used to localize a target, but a supine position triggers an increased reliance on visual cues \cite{oman_role_2003}, \cite{cheung_visually-induced_1990} and may prompt a compensatory neural response to overcome perceptual incongruence \cite{loued-khenissi_birds_2020}. Gravity is thought to emerge from an internal model that integrates multimodal inputs \cite{lacquaniti2014multisensory}. Visuo-proprioceptive processing relies partly on vestibular signals transmitted to the brain from the inner ear, providing information on a person's head orientation and movement in the earth's gravitational field \cite{bernard-espina_how_2022}. Gravity plays an anchoring role in human spatial perception by imposing a downward direction via the vestibular system \cite{clark_effects_2022}. Therefore, a vestibular system influenced by changes in gravity may induce perceptual alterations.

\subsubsection{Effects of altered gravity on humans}
Gravity is an embodied prior \cite{jorges_gravity_2017} whose absence has a known physiological impact on humans \cite{van_ombergen_effect_2017}. A paucity of research on behavioral consequences of hypo (and hyper) gravity persist \cite{mammarella_effect_2020}. Hypogravity alters the perception of written letters \cite{clement_mental_2009}, affects both horizontal and vertical distance perception \cite{clement_horizontal_2020}, and visual and haptic depth perception \cite{morfoisse_does_2020}. In real conditions of spaceflight, returning astronauts experience changes in vertical judgments \cite{harris_effect_2017} and studies on the effects of extended space missions have found impairments in visuomotor tasks \cite{manzey_mental_1998}. The gravity prior is thought to underlie an increased precision of downward versus upward moving objects \cite{torok_getting_2019}, sustained even when confronted with a “naturally” upward moving object, such as a rocket \cite{gallagher_gravity_2020}. Further studies have found that gravity alterations induce illusions of motion, underestimation of distance, delay in acquiring visual targets, and impairments in locomotion \cite{reschke_vestibular_2018}.Yet not all gravity-induced effects appear detrimental. Gravity facilitates movement planning \cite{saradjian_gravity-related_2014}, perspective-taking \cite{meirhaeghe_selective_2020}, \cite{Pfeiffer2018VestibularMO}, \cite{Pfeiffer2016VisualGC}, \cite{Pfeiffer2013MultisensoryOO} and shortens reaction times, in particular for complex tasks \cite{wollseiffen_physiology_2016}. Spaceflight induces  adaptive plasticity alongside neural dysfunction in human operators \cite{hupfeld_microgravity_2021}. Research is required to determine which dimensions of human performance are subject to either the deleterious or beneficial effects of altered gravity.

\subsubsection{Micro and hyper-gravity induced changes in uncertainty processing}
Few studies have explicitly looked at decision-making under uncertainty in altered gravity. One study found gravity effects in time-to-passage estimation \cite{indovina_anticipating_2013} where estimates of constant, versus gravity-congruent, vertical displacement were specifically less precise, suggesting a difficulty in integrating second-order uncertainty when estimating percepts outside of 1 G. Another study found that vestibular input dampens risk-taking \cite{de_maio_galvanic_2021}. In a similar vein, a supine position was found to decrease entropy in a random-number generation task \cite{gallagher_gravity_2019}. Ambiguous figures are ideal stimuli with which to assess uncertainty processing and have been used to probe microgravity-induced differences in perception. Bistable percepts were found to be more stable in microgravity, suggestive of uncertainty dampening \cite{clement_perceptual_2012}. In addition, a bias towards a given percept of an ambiguous figure, common in normogravity, disappears over time in orbit, suggesting that, perceptual uncertainty is weighted differently in microgravity \cite{clement_long-duration_2015}. Altogether, a precise understanding of variations in decision-making under uncertainty in altered gravity remains lacking. 

\subsubsection{Vision-based drone navigation}
Vision-based navigation of micro aerial vehicles enables robots to sense, plan, and execute flights based solely on onboard camera footage, without relying on global positioning information or other onboard sensors. This approach is particularly important in novel environments where global positioning information is unavailable. While autonomous vision-based navigation of micro aerial vehicles has recently achieved impressive results in agile flight \cite{Loquercio2021LearningHF}, swarm navigation \cite{Zhou2022SwarmOM}, and Mars surface exploration \cite{schroeder_nasas_2020}, \cite{witze_lift_2021}, there is a need to develop solutions for human-piloted or shared autonomy tasks \cite{Udupa2021SharedAI} where humans make important decisions in real-time. One of the main challenges for both expert human pilots \cite{Barin2017UnderstandingDP} and vision-based autonomous drones is  \emph{collision avoidance}\cite{Penicka2022MinimumTimeQW}. It is unknown how altered gravity conditions will impact operator performance in making accurate decisions on collision avoidance. Therefore, it is essential to investigate how altered gravity conditions might affect human operators self-motion perception during a vision-based autonomous flight. 

\subsubsection{Hypotheses}
The question of decision-making under uncertainty in human space exploration has been a fundamental concern from the earliest days of the human space program \cite{kanki2018cognitive}. The harrowing events of the Apollo 13 flight \cite{francis2020apollo} for instance highlighted the need to address the unexpected via engineering. This study queried how gravity alterations affect different facets of decision-making under uncertainty in a space analog environment (parabolic flight). We cast our experiment in a (partial) predictive coding framework to probe the role of stimulus uncertainty (formalized as entropy \cite{shannon2001mathematical}; \cite{loued-khenissi_information_2020}) on behavior. Using a novel self-motion estimation task in virtual reality, we tested the following set of confirmatory hypotheses: stimulus uncertainty prompts more incorrect responses, longer reaction times, and decreased subjective confidence reports. Our exploratory work included the following hypotheses: alterations in gravity induces changes in uncertainty processing observed in decision-making behavior including performance, reaction time and metacognition. 

\section{Results}

To test the effect of gravity alterations on decision making in drone navigation, two experiments were conducted, one in laboratory conditions in a sample of 22 subjects (on-ground experiment) and one in parabolic flight conditions in two subjects (parabolic flight experiment).

\subsection{On-ground experiment}

The purpose of the on-ground experiment was to develop an experimental paradigm to probe decision making under uncertainty in the context of drone navigation. The experiment served on the one hand to establish the stimulus material and task procedures for the parabolic flight experiments, in which a limited amount of task repetitions could be completed. On the other hand, the on-ground experiment served to collect data for baseline performance in a sufficiently large subject sample to be compared against the parabolic flight performance. A total of 22 subjects completed a Self-Motion Estimation task, where they were asked to watch a short video sequence showing a first-person camera view of a drone traversing a cave. Unbeknownst to the participant, the drone followed one of 21 possible trajectories differing in the angle at which the drone traversed the cave, of which 11 trajectories were associated with exiting the cave (drone trajectory angles 0–4, corresponding to pitch angles of 0\textdegree–7.83\textdegree) and 10 trajectories with a collision with the cave walls (drone trajectory angles 5–10, corresponding to pitch angles of 9.79\textdegree–19.58\textdegree). After watching the short movie clip, participants were asked to decide whether the drone would collide with the wall or exit the cave unscathed (2-alternative forced choice) and to then indicate their confidence (on a scale from 0=no confidence to 10=highest confidence). Each participant completed the task in Upright and Supine conditions.

\subsubsection{Task performance}

A total of 22 volunteers (6 female, median age: 25.6 years, age range: 22 - 43 years) were recruited for the study. 
Handedness was assessed with the Edinburgh Handedness Inventory \cite{Oldfield1971-vg}. Out of 22 subjects, 16 were right handed (Laterality Quotient between 61 and 100), 6 were ambidexterous (Laterality Quotient between -60 and 60), and 0 were left handed (Laterality Quotient between -100 and -61). 
All participants were healthy without prior history of psychological or neurological impairments, according to self-report. 
The study was conducted at the Robotics and Perception Group laboratory at the University of Zurich. Participants were asked to read and sign information and consent forms prior to the experiment. 
After arrival at the lab, the participant was equipped with VR goggles used for stimulus presentation and a controller for response collection.
After having received task instructions and a few training trials, the participants performed the experiment.
A total of 420 trials were performed across two conditions: sitting upright on a chair (Upright condition, 210 trials), and lying down on a camping mat (Supine condition, 210 trials). 
Each experimental condition was presented twice, resulting in a total of four blocks with 105 trials per block.
The order of the four experimental blocks was pseudo-randomized using either an ABBA or BAAB order (where A=Upright, B=Supine) to account for temporal effects on participants' performance, such as fatigue, habituation, or learning effects.
Participants were randomly assigned to the ABBA or BAAB order, such that half completed the experiment with ABBA, and the other half with the BAAB order.
The Supine condition served as a rudimentary proxy of a gravity-free condition (simulated microgravity). 
In between experimental blocks, participants completed a brief questionnaire (NASA TLX, \cite{hart_tlx_1986,hart_tlx_2006}) to index their subjective assessment of task difficulty.
The experiment lasted approximately 1 hour in total and conformed to the University of Zurich ethics guidelines.

\begin{figure}[!ht]
\centering
\begin{center}
\includegraphics[trim = 0mm 0mm 0mm 0mm, clip, width=.49\linewidth]{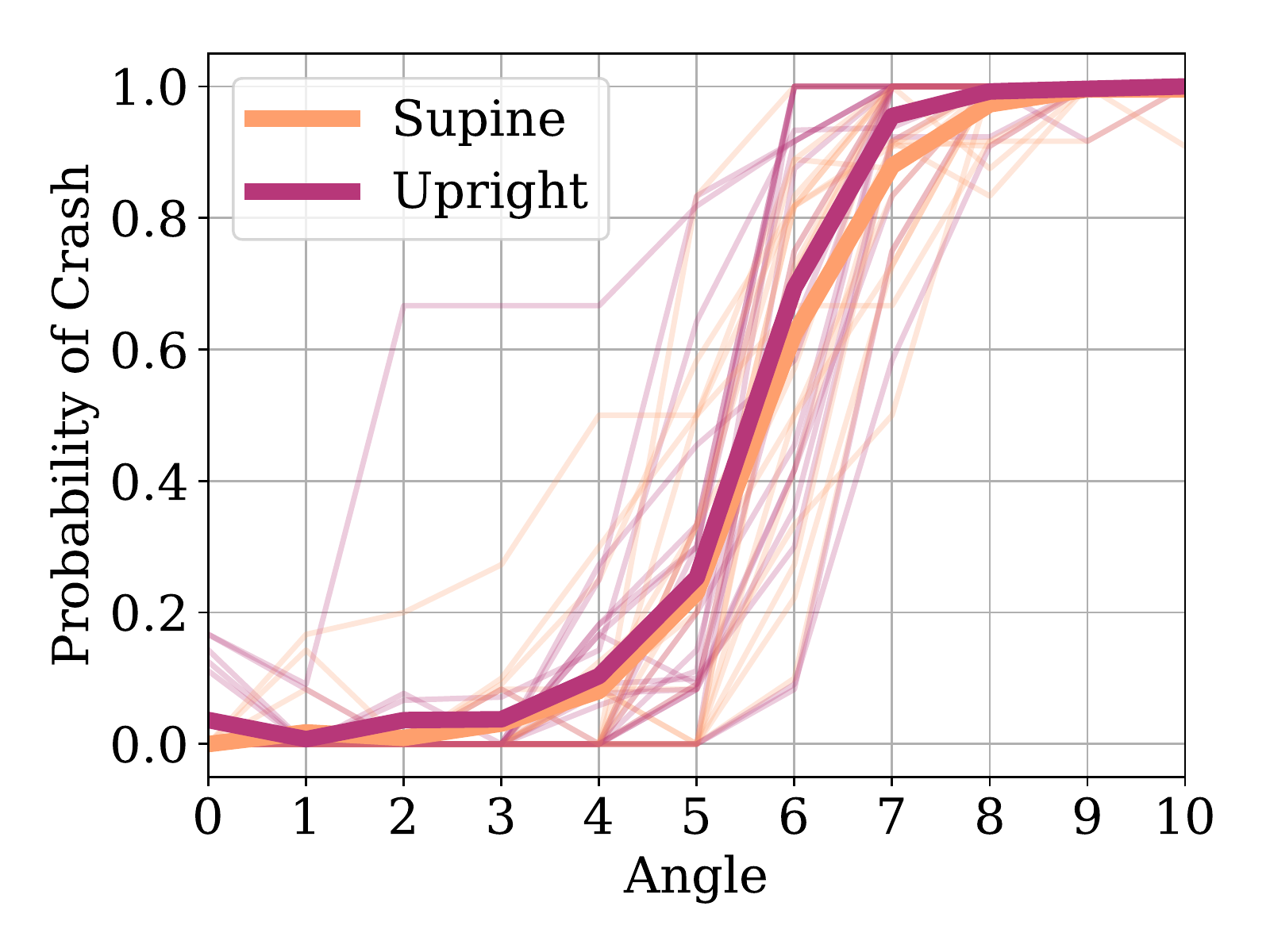} 
\includegraphics[trim = 0mm 0mm 0mm 0mm, clip, width=.49\linewidth]{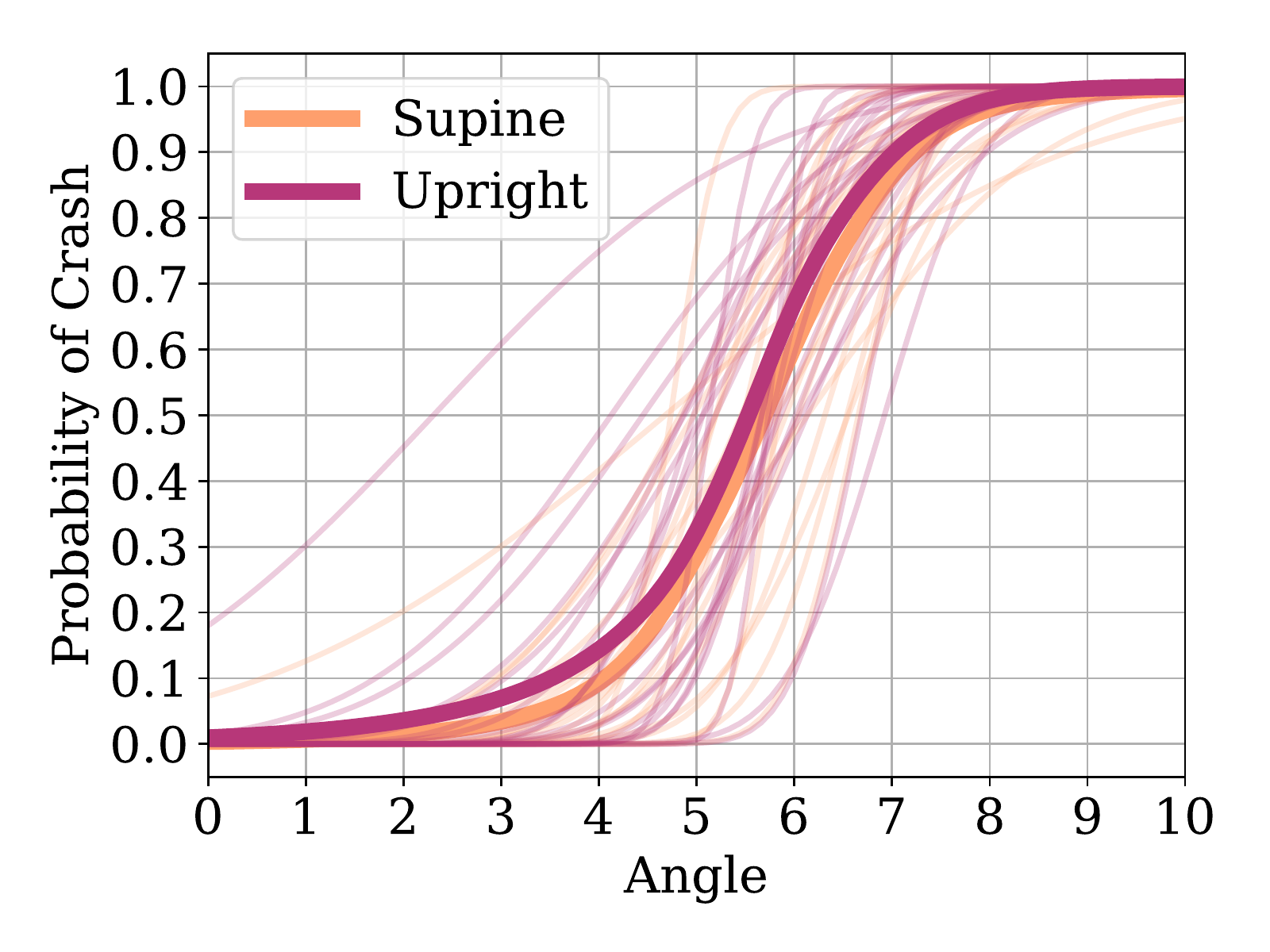}  
\caption[caption]{On-ground experiment individual subject's average collision responses (left panel) and psychometric function fits on the same data (right panel) are shown as functions of drone trajectory angle and gravity condition. Thin lines show individual subject data, and thick lines show group averages.}
\vskip -0.3in 
\label{fig:onground-collision-response}
\end{center}
\end{figure}

\begin{figure}[!ht]
\centering
\begin{center}
\includegraphics[trim = 0mm 0mm 0mm 0mm, clip, width=.49\linewidth]{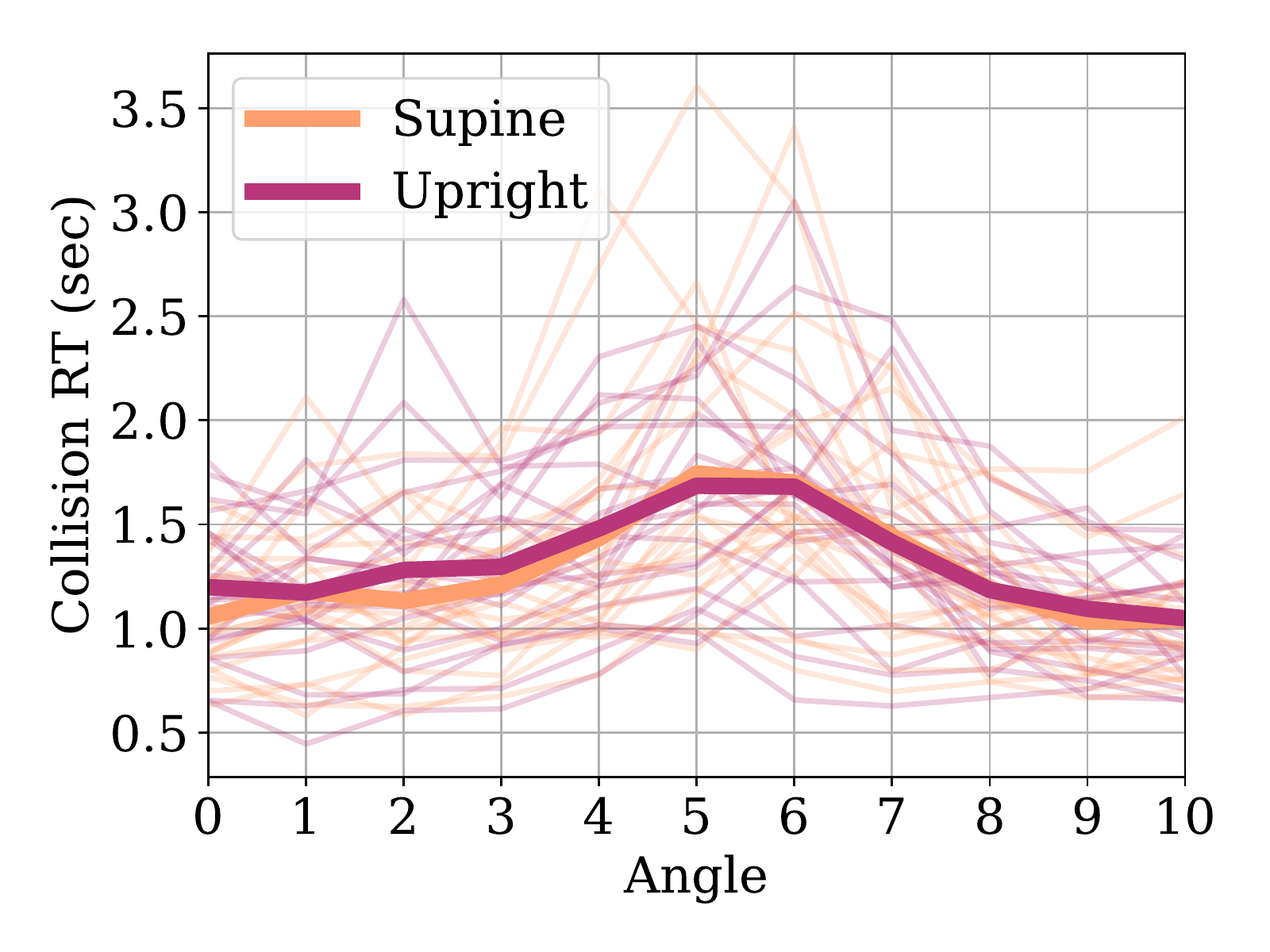} 
\includegraphics[trim = 0mm 0mm 0mm 0mm, clip, width=.49\linewidth]{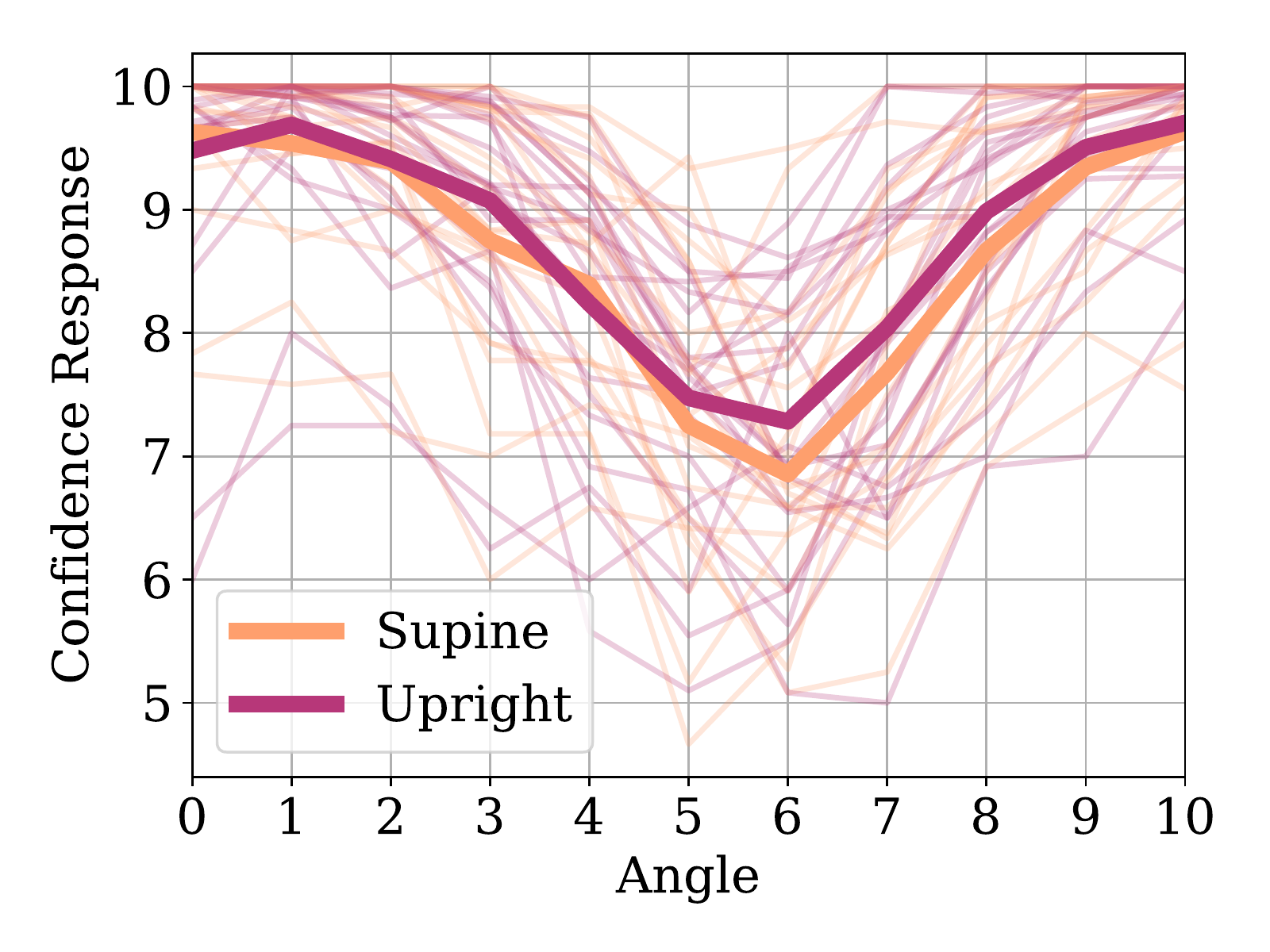}  
\caption[caption]{On-ground experiment collision response times (left panel) and confidence responses (right panel) are shown as functions of drone trajectory angle and gravity condition. Thin lines show individual subject data, and thick lines show group averages.}
\vskip -0.3in 
\label{fig:onground-confidence-rating}
\end{center}
\end{figure}

\begin{figure}[!ht]
\centering
\begin{center}
\includegraphics[trim = 0mm 0mm 0mm 0mm, clip, width=.49\linewidth]{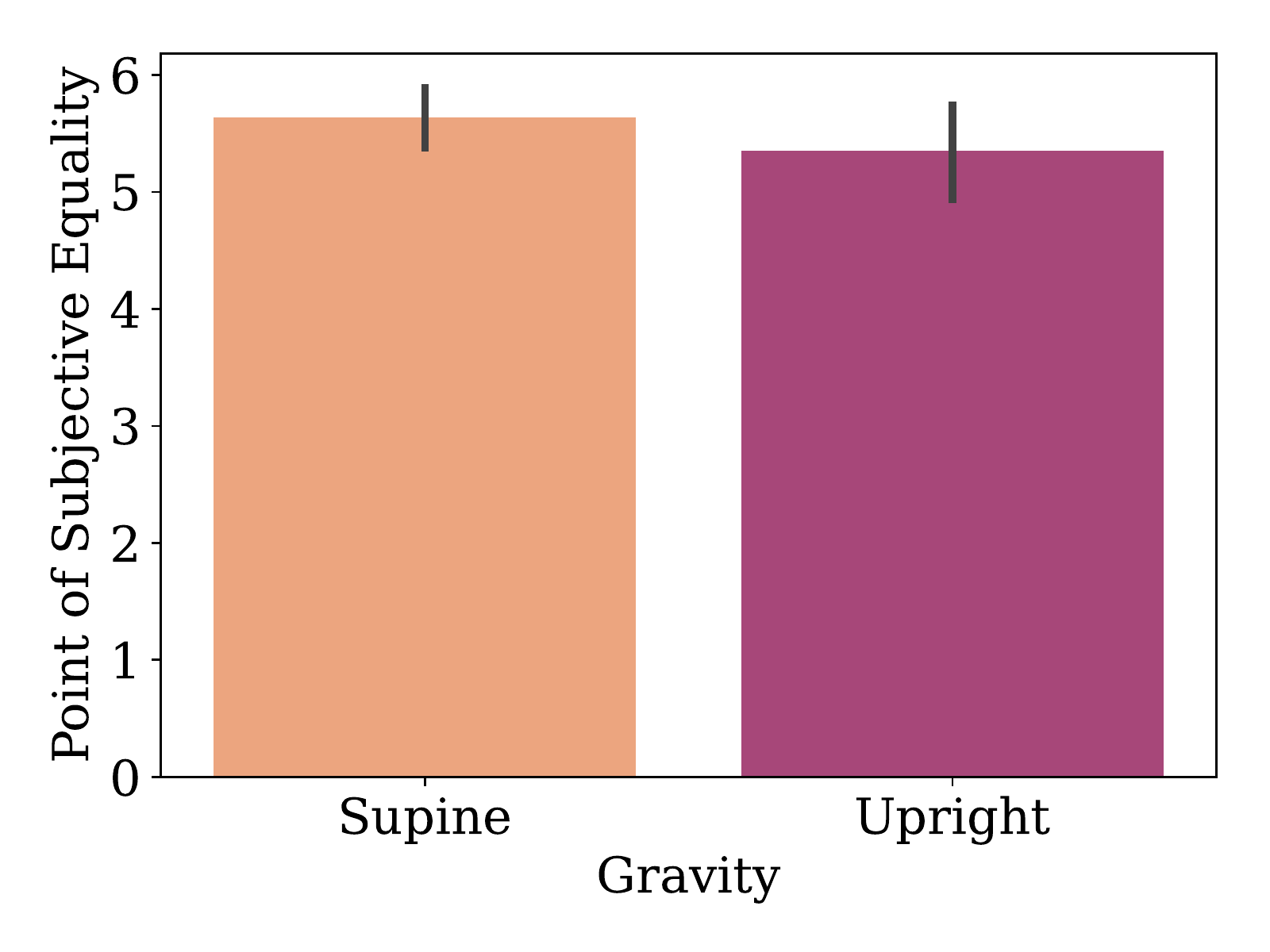}  
\includegraphics[trim = 0mm 0mm 0mm 0mm, clip, width=.49\linewidth]{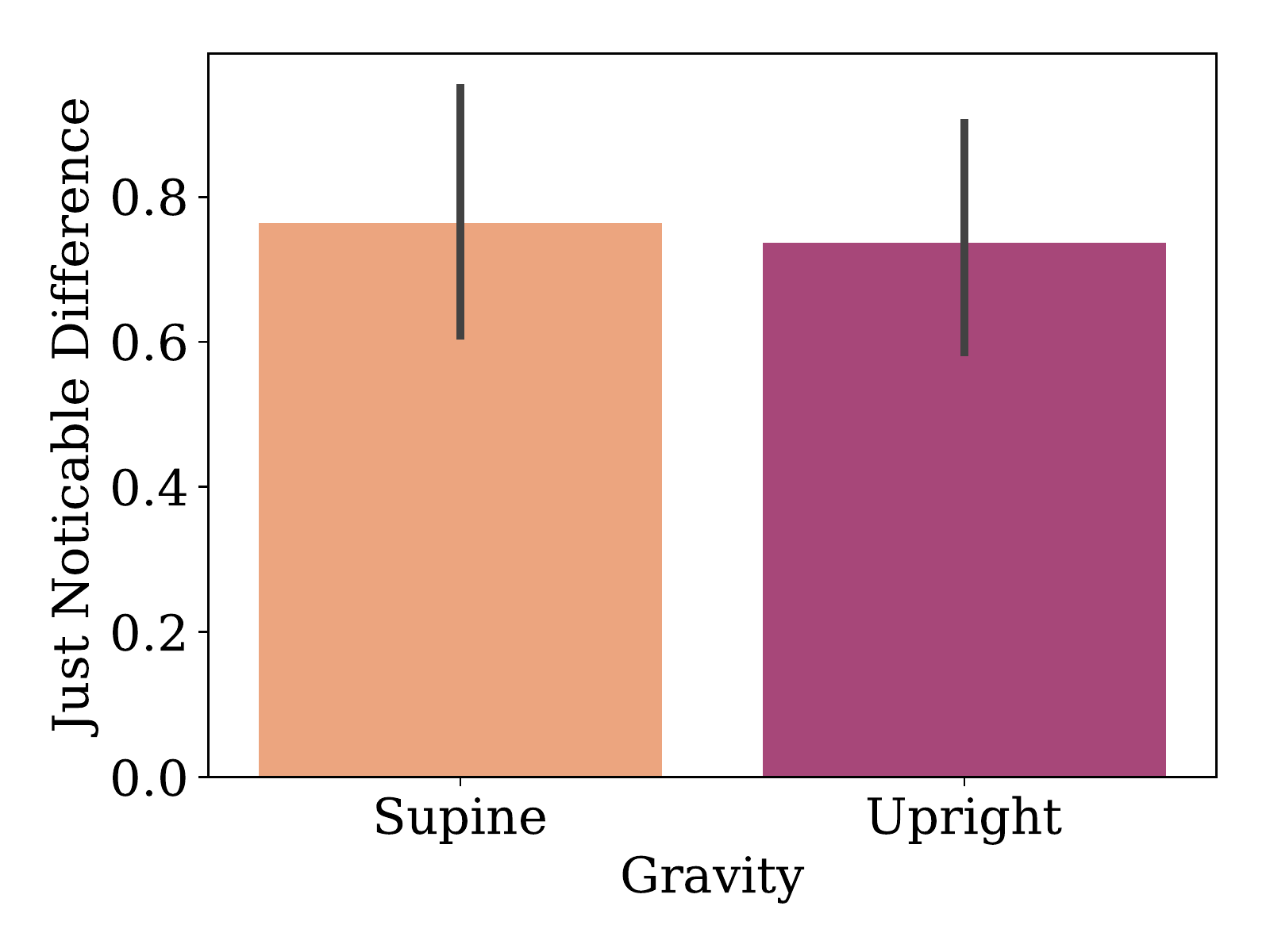}  
\caption[caption]{On-ground experiment psychometric fit results for collision responses showing the group average Point of Subjective Equality (left panel) and Just Noticeable Difference (right panel) for different gravity conditions. Error bars indicate the Standard Error of the Mean.}
\vskip -0.3in 
\label{fig:onground-psychometric-fitting-results}
\end{center}
\end{figure}

\begin{figure}[!ht]
\centering
\begin{center}
\includegraphics[trim = 0mm 20mm 0mm 20mm, clip, width=.99\linewidth]{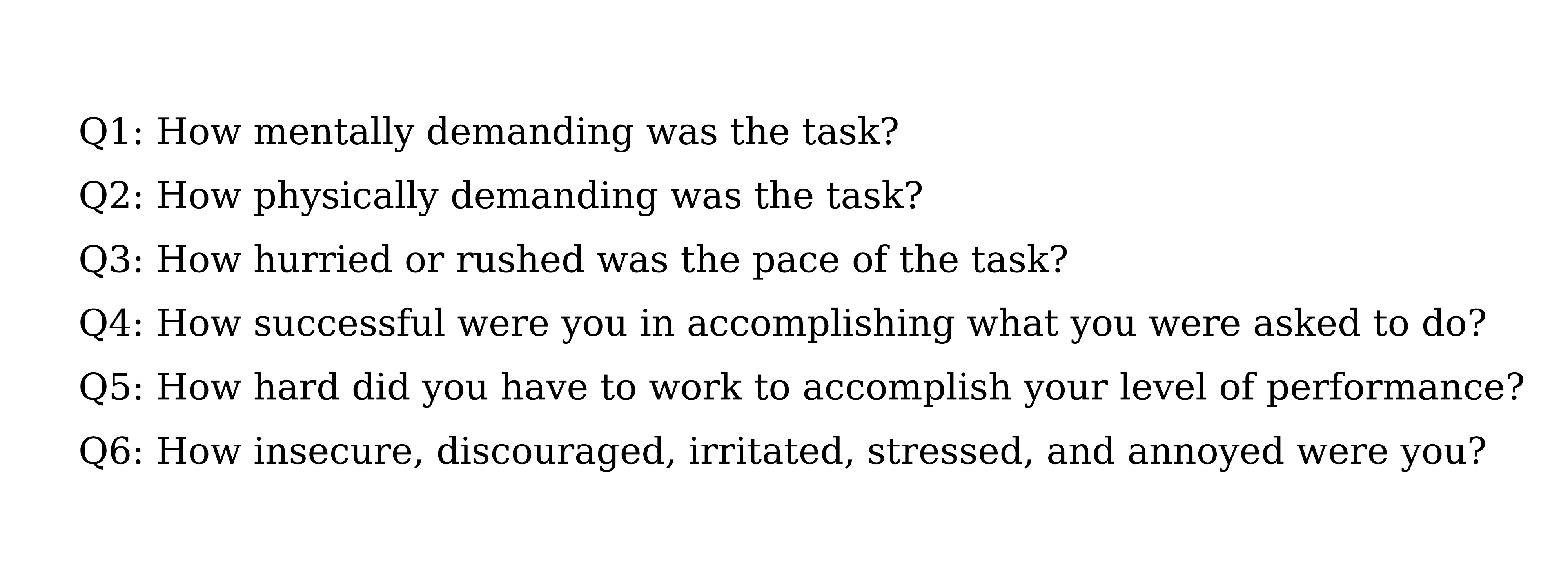} 
\includegraphics[trim = 0mm 0mm 0mm 0mm, clip, width=.99\linewidth]{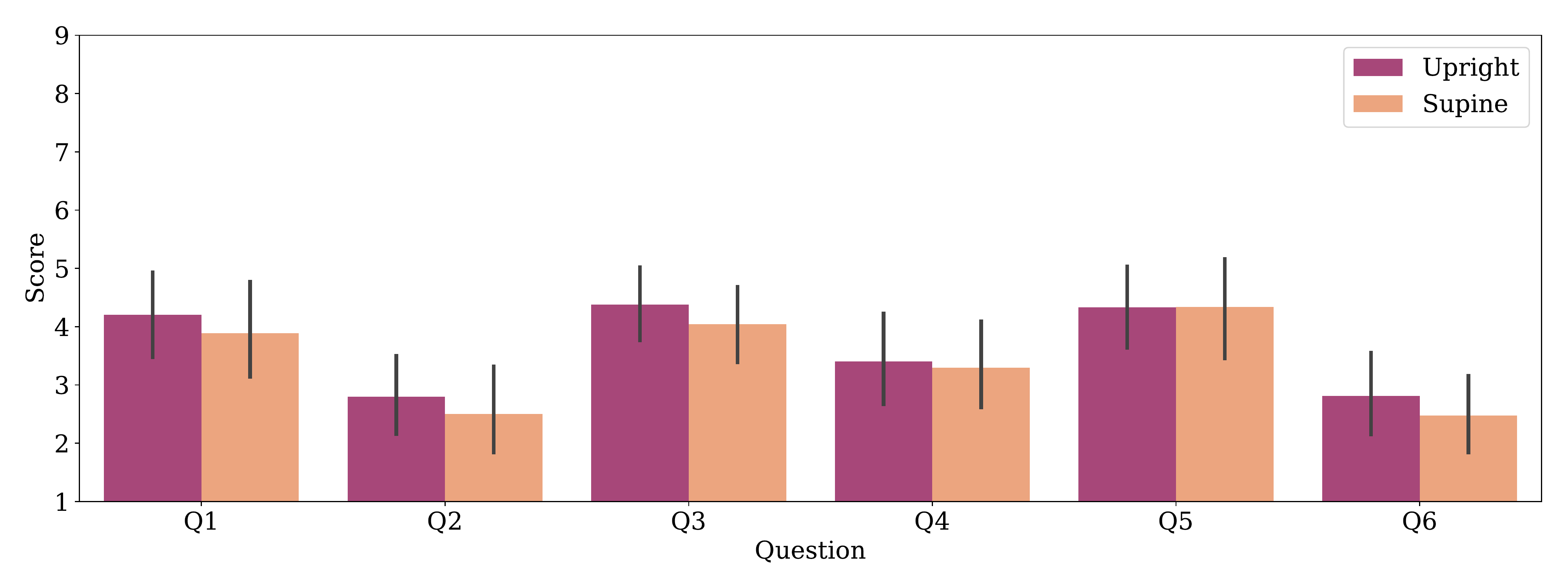} 
\caption[caption]{On-ground experiment NASA TLX questionnaire ratings on a scale from 1 (Very Low) to 9 (Very High). Bars show average ratings and error bars show the standard error of the mean.}
\vskip -0.3in 
\label{fig:onground-tlx}
\end{center}
\end{figure}

Figures \ref{fig:onground-collision-response} and \ref{fig:onground-confidence-rating} show the grand average collision ratings and response times. Our experimental task induced a systematic modulation of the collision probability ratings, reflected in a gradual increase of collision rating probability between drone trajectory angle 4 (pitch angle of 7.83\textdegree) and 7 (pitch angle of 13.71\textdegree). At the same time, our experimental paradigm induced a modulation of the decision uncertainty, reflected in longer response times for collision ratings.

Out of 22 participants, the data of one participant was excluded due to a technical problem with the response box, where ratings were not recorded. Subsequent analyses were conducted on the data of the remaining 21 participants. A chi-square test on collision responses shows a significant tendency to select no-crash over crash responses in both Supine ($\chi^2$ = 9.62, p = 0.002) and Upright ($\chi^2$ = 32.23, p $<$ 0.001) conditions, but considerably more so in the Upright condition. 
Collision response time distributions for both conditions conform to an expected gamma distribution, while confidence responses do not. 
Average confidence ratings across conditions were similar overall (8.73, sd = +/- 1.66 and 8.89, sd = +/- 1.66,  in the Upright and Supine conditions, respectively). 
Overall, participants’ accuracy rate was 92.27\% for the Upright condition against 93.29\% in the Supine condition. 

\subsubsection{Psychometric analysis}

Psychometric analysis aims to quantitatively describe and compare decision-making under uncertainty between different gravity conditions. We fit psychometric functions on the collision response data of individual subjects. 
The first parameter we estimate is the Point of Subjective Equality (PSE), which indicates the drone trajectory angle at which participants assign an equal probability of crash vs. no-crash ratings (Figure 4). 
The second parameter is the Just Noticeable Difference (JND), which is the amount of change in drone trajectory angle required to induce a noticeable difference in subjective collision response ratings (Figure 4).
Psychometric analysis is carried out for each of the 21 subjects and two experimental conditions of Gravity modulation (levels: Upright, Supine). 
Psychometric fits are performed using Bayesian optimization \cite{Okamoto2012-mx}.
Figure \ref{fig:onground-psychometric-fitting-results} shows the group average PSE and JND parameter estimates from the on-ground experiment. Repeated-measures Analysis-Of-Variance (rmANOVA) for PSE and JND estimates, comparing the Upright and Supine conditions showed no significant differences (all p-values $>$ 0.05).

\subsubsection{NASA TLX}
The NASA TLX questionnaire was used to measure mental and physical demand related to the self-motion estimation task. Figure \ref{fig:onground-tlx} shows that overall participants gave average ratings, indicating that the task was not experienced as particularly stressful. Paired-samples t-tests showed no significant differences in ratings between the Upright and Supine conditions (all p-values $>$ 0.05).


\subsubsection{General linear mixed model analysis}

To look for differences in uncertainty processing across gravity conditions, we created 4 models, three for the first decision (collision prediction), and one for the second (confidence response). For each decision, outcome variables included participant response (correct response, collision prediction response or confidence estimate), and collision prediction response reaction times. 
 
At timepoint 1, we examined correct responses, collision response reaction time, and collision esponses, as a function of expected value of stimulus; stimulus uncertainty; and gravity condition (Model 1a-c, Table 1 \ref{table:regression-models}). Models included a random effects regressor to control for subject variability (n = 21, 4941 observations). Key predictors of interest are stimulus uncertainty and gravity condition, with the expected value included in models primarily as a means of controlling for the effect of prediction alone, but also as a means of verifying model assumptions.

\begin{table*}[ht!]
\centering
\begin{tabular}{||c c c c c||} 
 \hline
Model & Outcome Variable & Predictors & Random Effect  & Fit
 \\ [0.5ex] 
 \hline\hline
 1a & Correct Response & Stimulus Uncertainty, Gravity condition & Subject (on-ground experiment) & Logistic \\
 1b & Collision Prediction RT & Stimulus Uncertainty, Gravity condition & Subject (on-ground experiment) & Linear \\
 1c & Collision Prediction & Stimulus Uncertainty, Gravity condition, Response RT & Subject (on-ground experiment) &  Logistic \\
 2 & Confidence Response & Stimulus Uncertainty, Gravity condition, Confidence RT, correct response & Subject (on-ground experiment) & Linear \\
 \hline
\end{tabular}
\caption{Regression Models used for statistical analysis.}
\label{table:regression-models}
\end{table*}

\subsection{Model 1a}
Uncertainty alone predicted more incorrect responses  (z = -15.330, p $<$0.001) in line with confirmatory hypotheses. A higher expected value of crash also predicted a correct response (z = 7.586, p $<$ 0.001).

\subsection{Model 1b}
Uncertainty alone predicted increased response times  (z = 8.421, p $<$0.001) underlining the behavioral response to an ambiguous stimulus, as did Supine (z = 2.243, p = 0.025).

\subsection{Model 1c}
Stimulus value correlated with a positive collision response, (z = 35.916, p $<$ 0.001) but stimulus uncertainty led to a no-crash prediction (z = -5.426, p $<$ 0.001). However, positive collision responses were also predicted by the interaction between Supine and stimulus uncertainty (z = 3.367, p $<$ 0.001). Taken together, results suggest that uncertainty interacts with the Supine condition to alter perceptual decisions, in this case favoring a crash response. Interestingly, the main effect of stimulus uncertainty alone predicted a no-crash response, an unexpected finding.


\subsection{Model 2}
Stimulus uncertainty negatively affected confidence ratings (z = -7.544, p $<$ 0.001). In addition, increased stimulus uncertainty in the Supine condition correlated with lower confidence responses (z = -3.536, p $<$0.001). The association between stimulus uncertainty and lower confidence ratings cements the link between a covert, stimulus uncertainty and overt, declarative uncertainty. 

\subsubsection{Summary}

The on-ground experiment was performed to establish several theoretical and empirical baselines. In a first instance, the task presented was novel and meant to operationalize a graded perceptual uncertainty. This aim was accomplished based on the psychometric fits yielding sigmoidal curves across subjects. Second, we sought to examine if differences in body position (Upright and Supine) would yield differences in responses linked to decision-making and associated variables (reaction time, uncertainty and confidence). Although some differences emerged, it is not clear if these are due to simple differences in body position or as a proxy of microgravity. Third, the on-ground experiment acted as a validation of expected results, namely positive correlations between a latent, uncertainty and correct responses, as well as a negative associations with subjective, declared confidence responses. Finally and most importantly, the on-ground experiment yielded a baseline psychometric function for the two subjects slated to fly and perform a version of the study on-board a parabolic flight (Figure 7).

\subsection{Parabolic flight experiment}

\subsubsection{Task performance}

Unlike the on-ground experiment, chi-square tests on collision responses show an even distribution between positive and negative crash predictions (no significant difference in frequency) in all gravity conditions. 
Collision and confidence response time distributions for all gravity conditions follow an expected gamma distribution. 
Average confidence ratings were 7.3, sd = +/- 2.17 in normal gravity; 7.7, sd = +/- 1.77 in hypergravity, and 8.1, sd = +/- 1.52 in microgravity. 
Accuracy rates in flight were 63.6\%, 75\% and 68\% for normal, hyper- and microgravity, respectively. 

\subsubsection{Psychometric analysis}

Psychometric analysis for the parabolic flight data was conducted. 
Because of the low number of trials in the microgravity and hypergravity phase, we pooled the data across the two experiment participants and across the drone direction factor. 
Then, psychometric functions were fit to collision responses for each of the conditions of the 3 Gravity conditions (normogravity, microgravity, hypergravity) using bayesian optimization \cite{Okamoto2012-mx}.
Figure 5 \ref{fig:parabolic-psychometric-fitting-results} shows the psychometric function fit results. 
Inspection of the psychometric fit results indicate an effect of gravity on collision detection threshold and confidence ratings. 
Figure 5 shows that microgravity increased the collision detection thresholds (PSEs) and confidence ratings. 
Thus, under microgravity, drone collisions were noticed at a higher and with higher confidence than under normogravity and hypergravity conditions.

\begin{figure}[!ht]
\centering
\begin{center}
\includegraphics[trim = 0mm 0mm 0mm 0mm, clip, width=.49\linewidth]{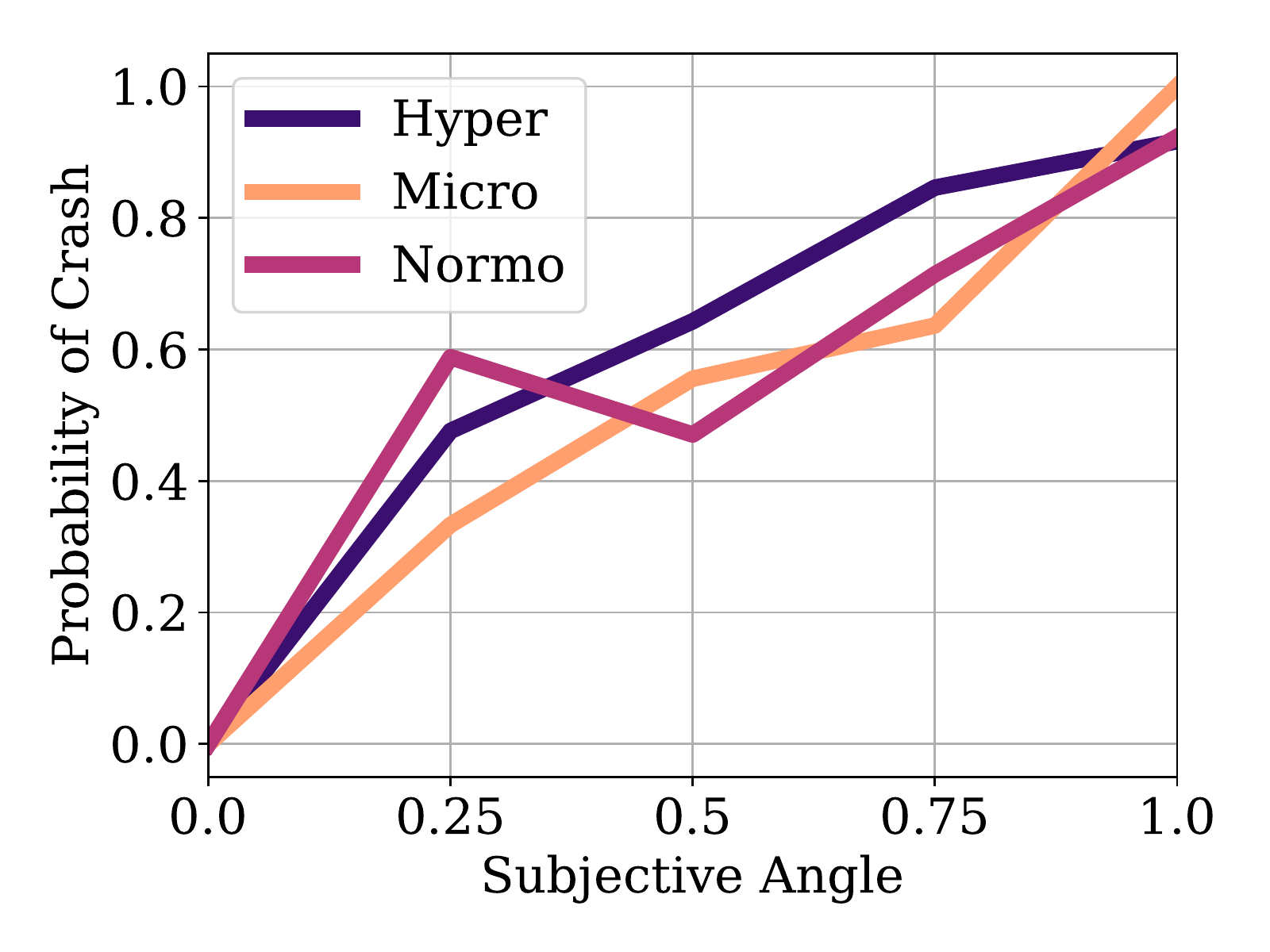}  
\includegraphics[trim = 0mm 0mm 0mm 0mm, clip, width=.49\linewidth]{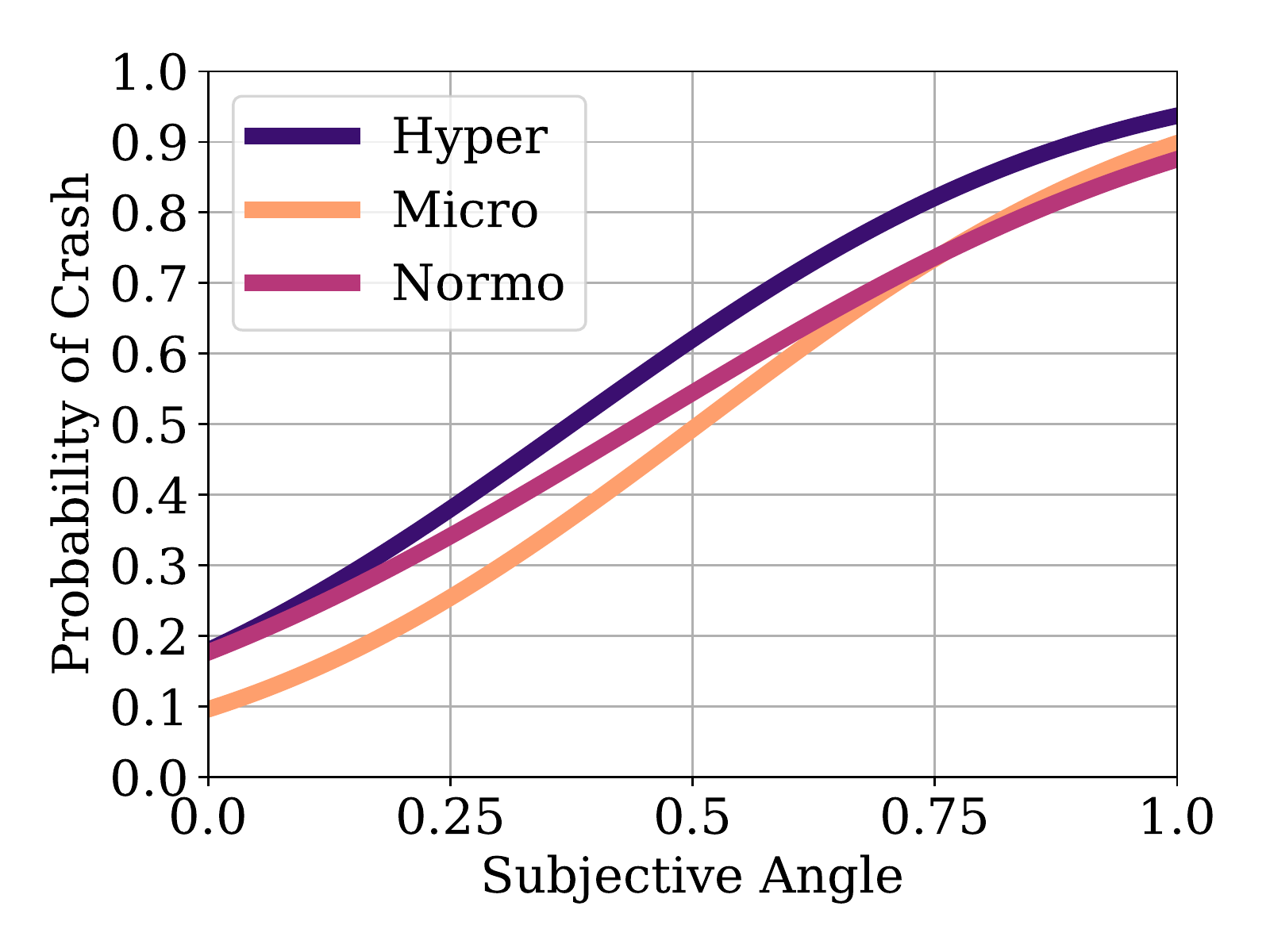}  
\caption[caption]{Parabolic Flight experiment collision response probability (left panel) and psychometric function fits (right panel) are shown as functions of Subjective Angle and gravity condition.}
\vskip -0.3in 
\label{fig:parabolic-collision-response}
\end{center}
\end{figure}

\begin{figure}[!ht]
\centering
\begin{center}
\includegraphics[trim = 0mm 0mm 0mm 0mm, clip, width=.49\linewidth]{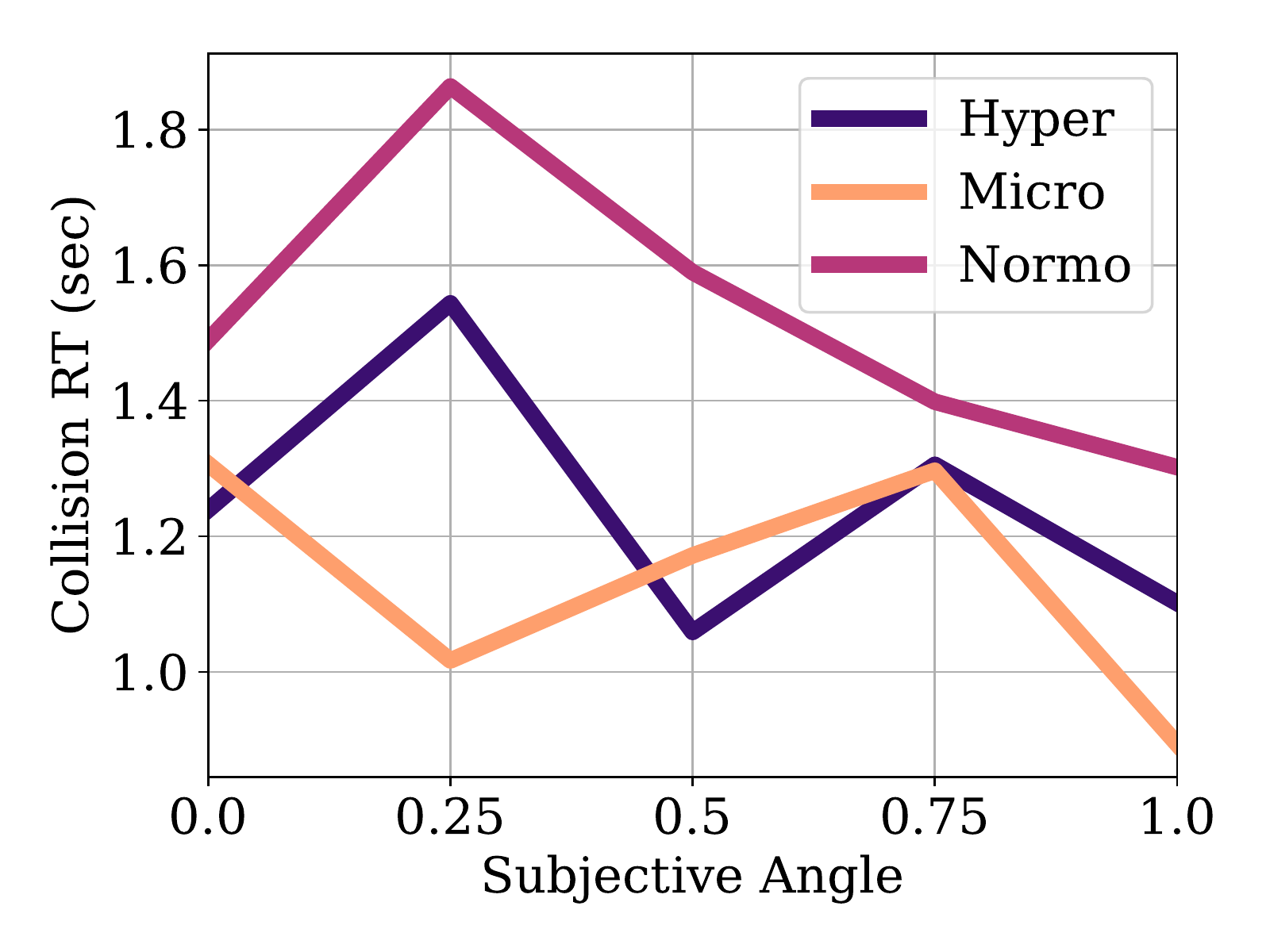} 
\includegraphics[trim = 0mm 0mm 0mm 0mm, clip, width=.49\linewidth]{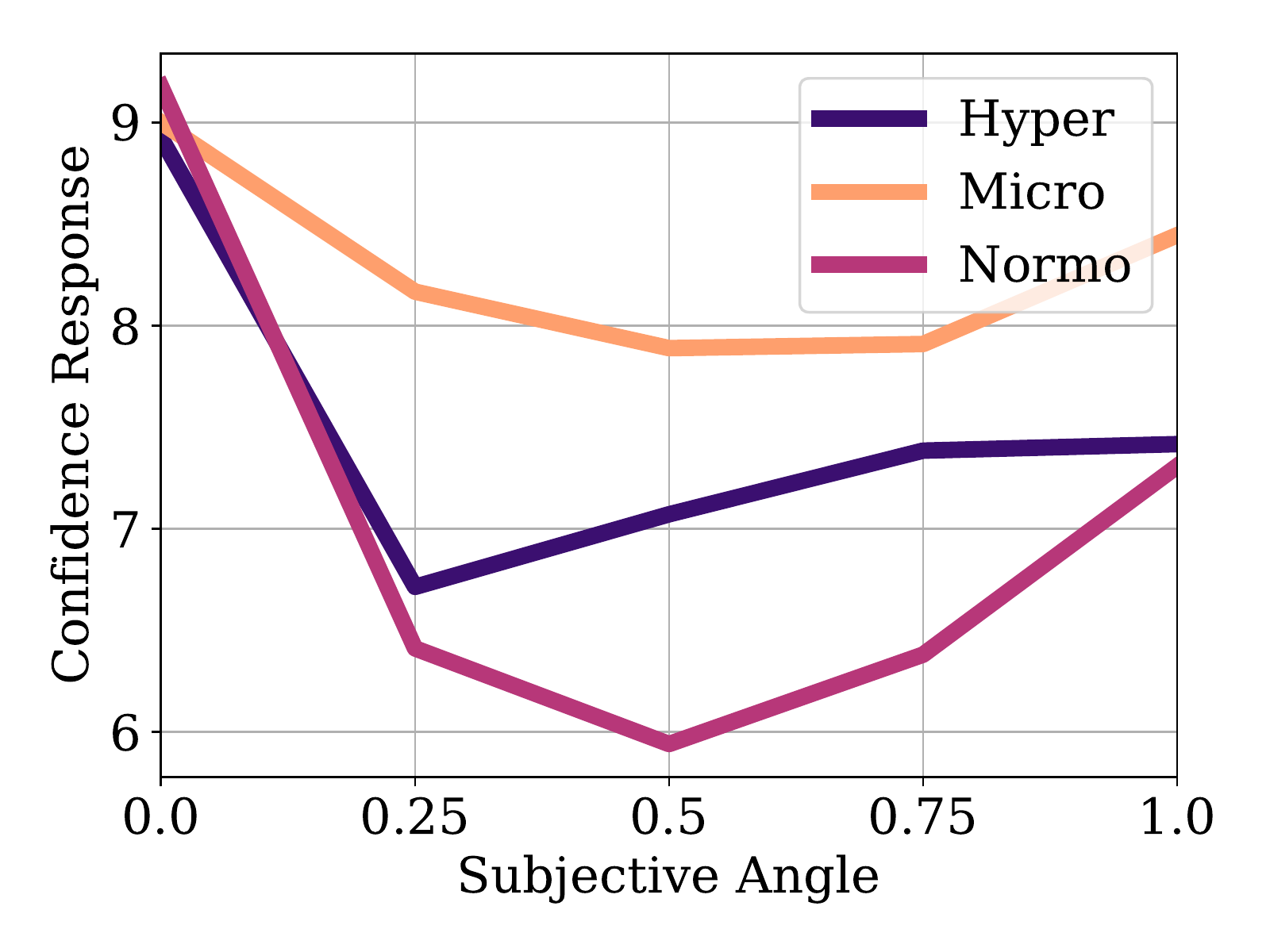}  
\caption[caption]{Parabolic Flight experiment collision response times (left panel) and confidence responses (right panel) are shown as functions of Subjective Angle and gravity condition.}
\vskip -0.3in 
\label{fig:parabolic-confidence-response}
\end{center}
\end{figure}

\begin{figure}[!ht]
\centering
\begin{center}
\includegraphics[trim = 0mm 0mm 0mm 0mm, clip, width=.49\linewidth]{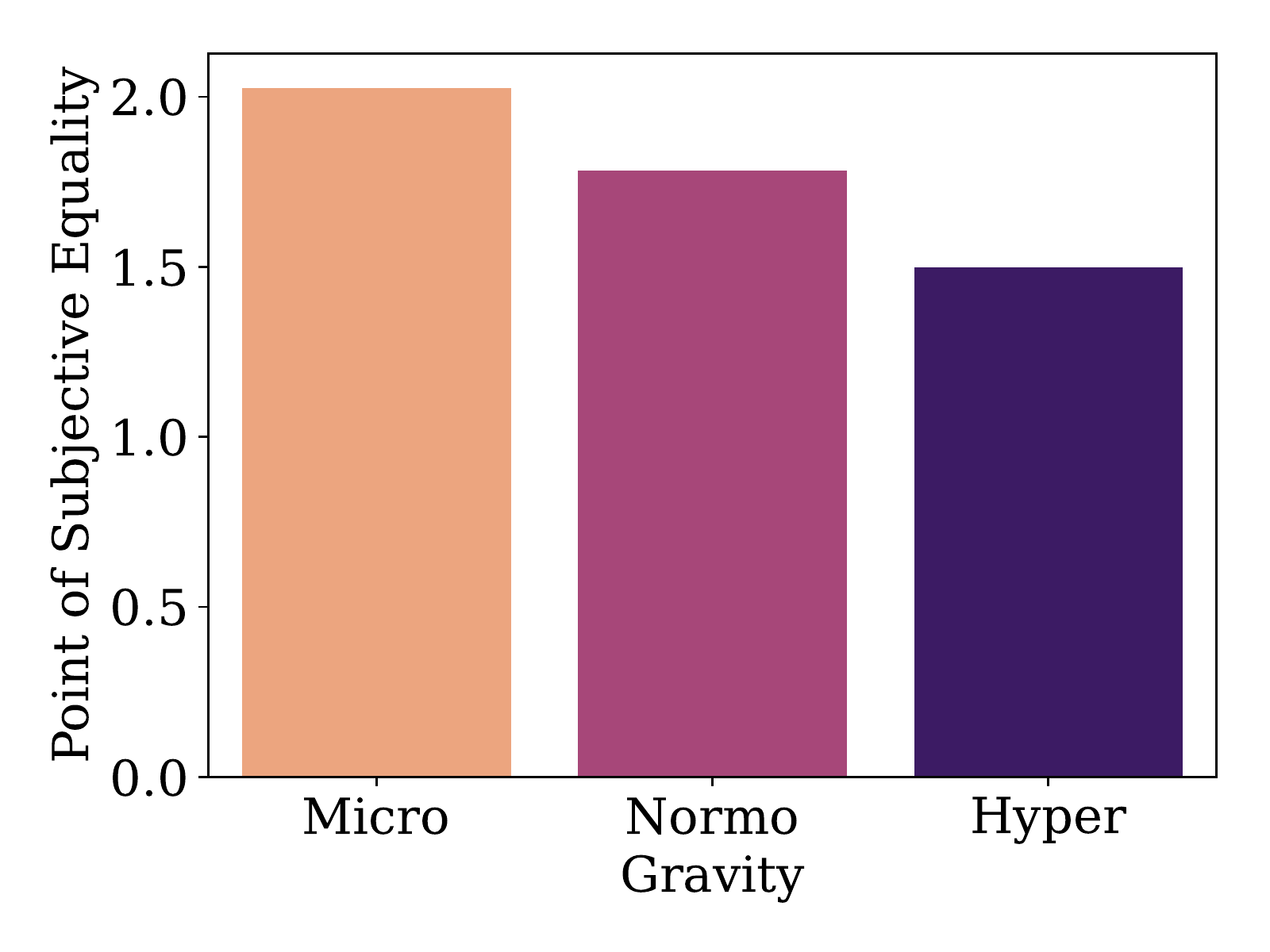}  
\includegraphics[trim = 0mm 0mm 0mm 0mm, clip, width=.49\linewidth]{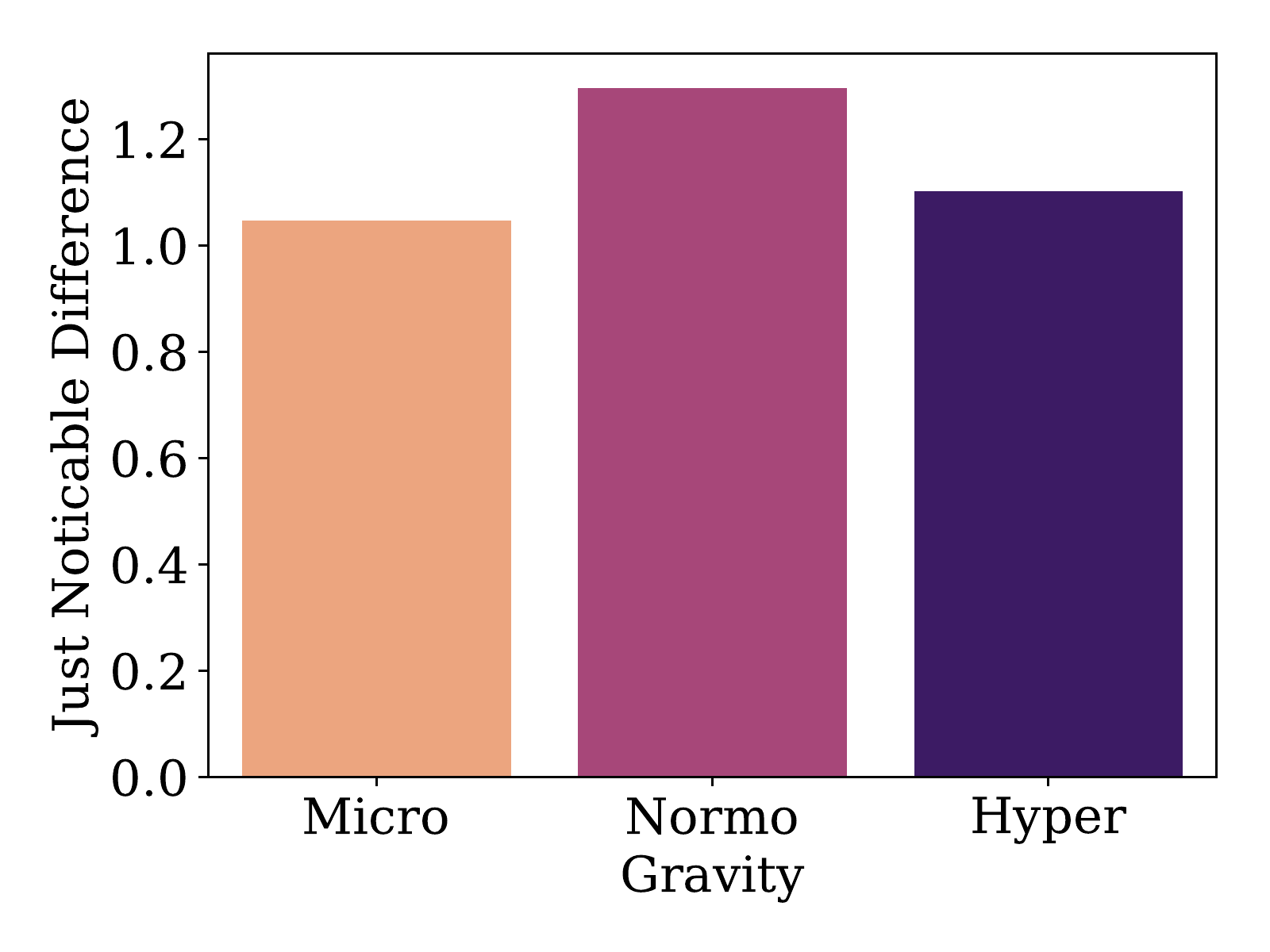}  
\caption[caption]{Collision response psychometric fit parameters Point of Subjective Equality (PSE, left panel) and Just Noticeable Difference (JND, right panel) for the different gravity conditions of the Parabolic Flight experiment.}
\vskip -0.3in 
\label{fig:parabolic-psychometric-fitting-results}
\end{center}
\end{figure}

\subsubsection{Regression analysis}

Similar to the on-ground experiment, we detailed four models to examine perceptual decision-making in-flight, one examining correct collision response, collision response RT, collision response, and confidence ratings as a function of stimulus uncertainty in interaction with gravity conditions as well as the average value of the stimulus. 

\subsection{Model 1a}
Longer response times were associated with correct collision response (z = 2.114, p = 0.034) as was the expected value of the stimulus (z = 2.960, p = 0.003) while uncertainty correlated positively with incorrect responses (z = -7.843,  p $<$0.001) in line with confirmatory hypotheses. 

\subsection{Model 1b} 
There was no effect of independent variables on collision response times. 

\subsection{Model 1c}
Increased stimulus uncertainty predicted a crash response (z = 3.305, p = 0.001) in contrast to the on-ground experiment, but in line with expected findings. Expected value of the stimulus, based on the on-ground experiment data, also positively predicted a crash response (z = 10.792, p $<$ 0.001). In addition, the hypergravity condition marginally predicted a crash response, though this effect did not reach significance (z = 1.920, p = 0.055).


\subsection{Model 2}
Confidence ratings were significantly higher under conditions of microgravity (z = 2.540, p = 0.011). However, while stimulus uncertainty significantly predicted low confidence scores ( z = -11.256, p $<$0.001) in line with our confirmatory hypothesis, stimulus uncertainty under conditions of microgravity led to higher confidence ratings (z = 1.968, p = 0.049). This finding, though marginal, encapsulates the potential impact of hidden variables on decision-making in altered gravity, in this case on the metacognitive aspect of performance. 

\subsection{Summary}
In-flight performance yielded several insights into decision-making under uncertainty across gravity conditions. Of note, while absolute performance deteriorated relative to the on-ground experiment, we found no significant differences linked to gravity conditions. Stimulus uncertainty broadly affected participants’ responses in expected ways but the differential effect of uncertainty in microgravity on overconfidence offers insight into performance deviations in space flight (Figure 7). 

\begin{figure}
\begin{multicols}{2}
    \centering
    \text{On-Ground Experiment\newline\newline}\par
    \includegraphics[width=\linewidth]{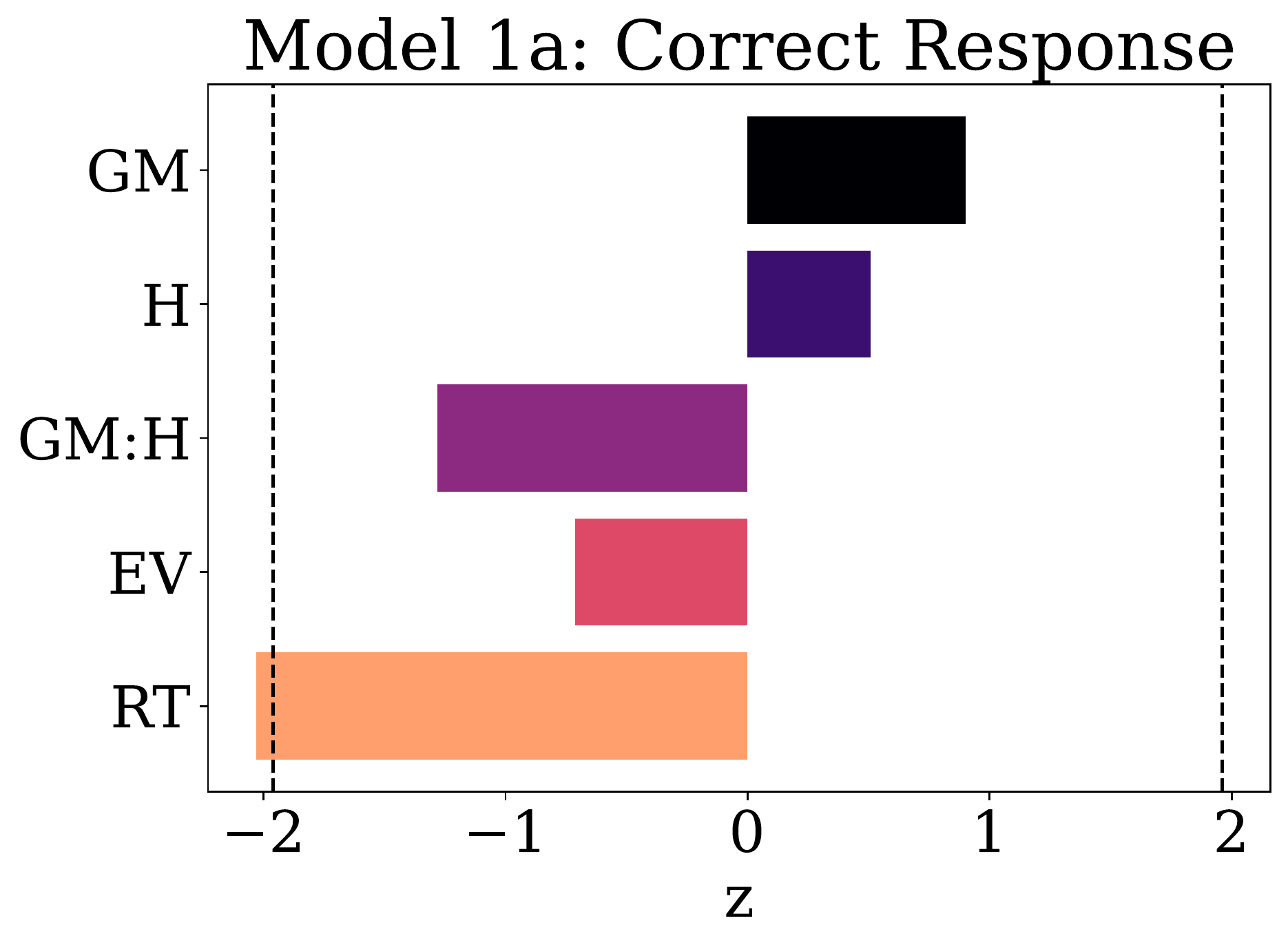}\par
    \text{Parabolic Flight Experiment\newline\newline}\par
    \includegraphics[width=\linewidth]{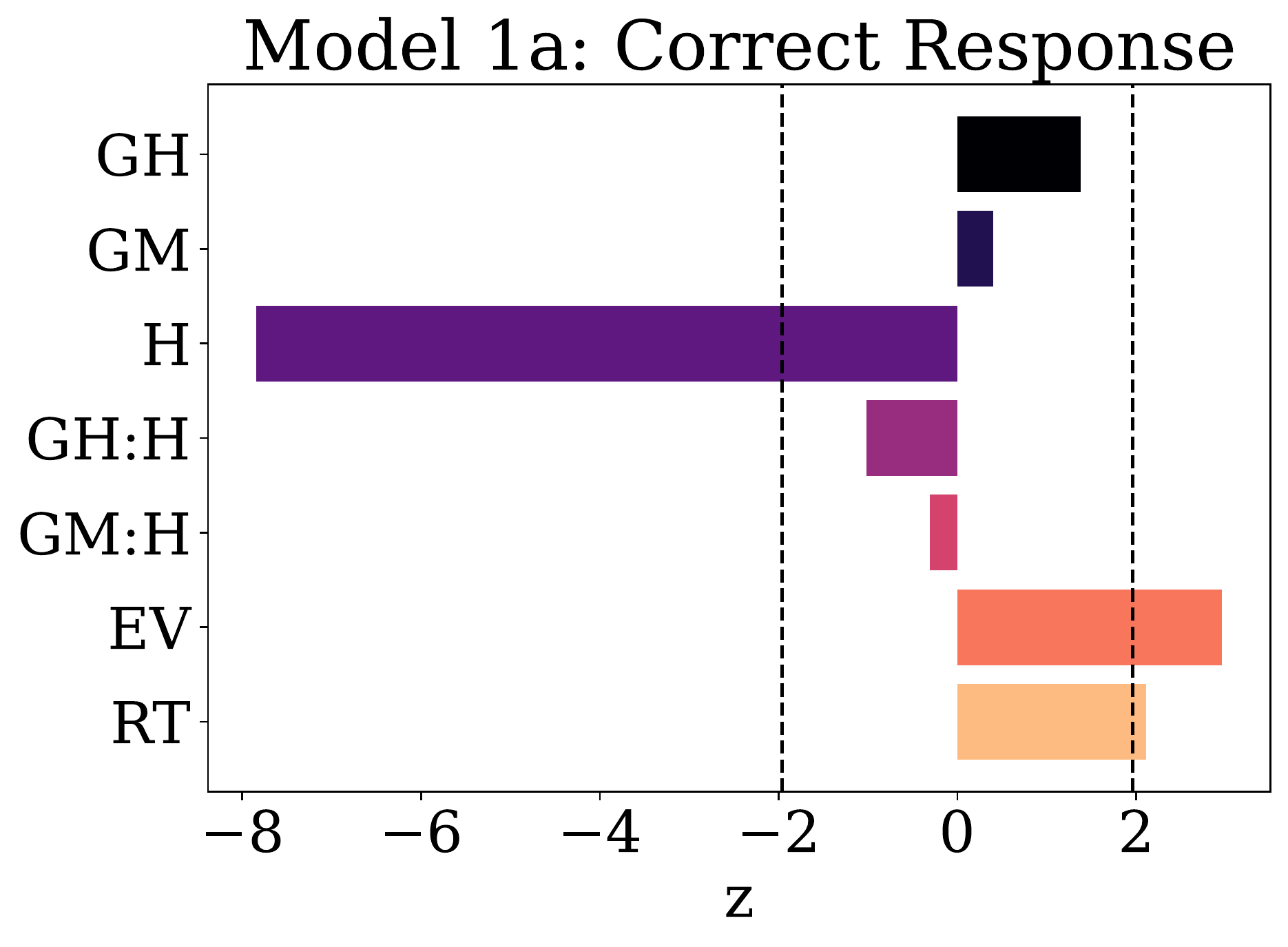}\par 
\end{multicols}
\begin{multicols}{2}
    \centering
    \includegraphics[width=\linewidth]{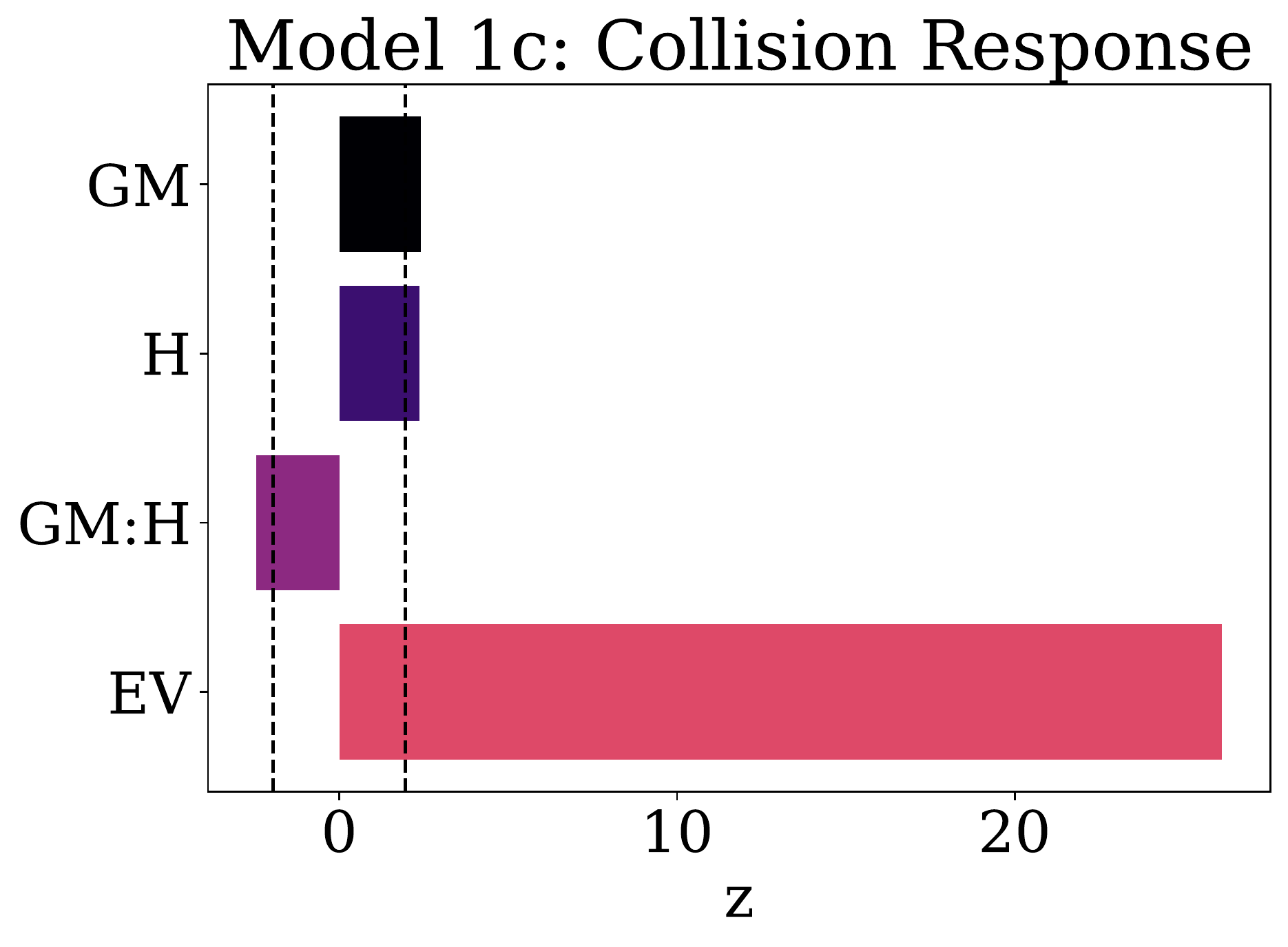}\par
    \includegraphics[width=\linewidth]{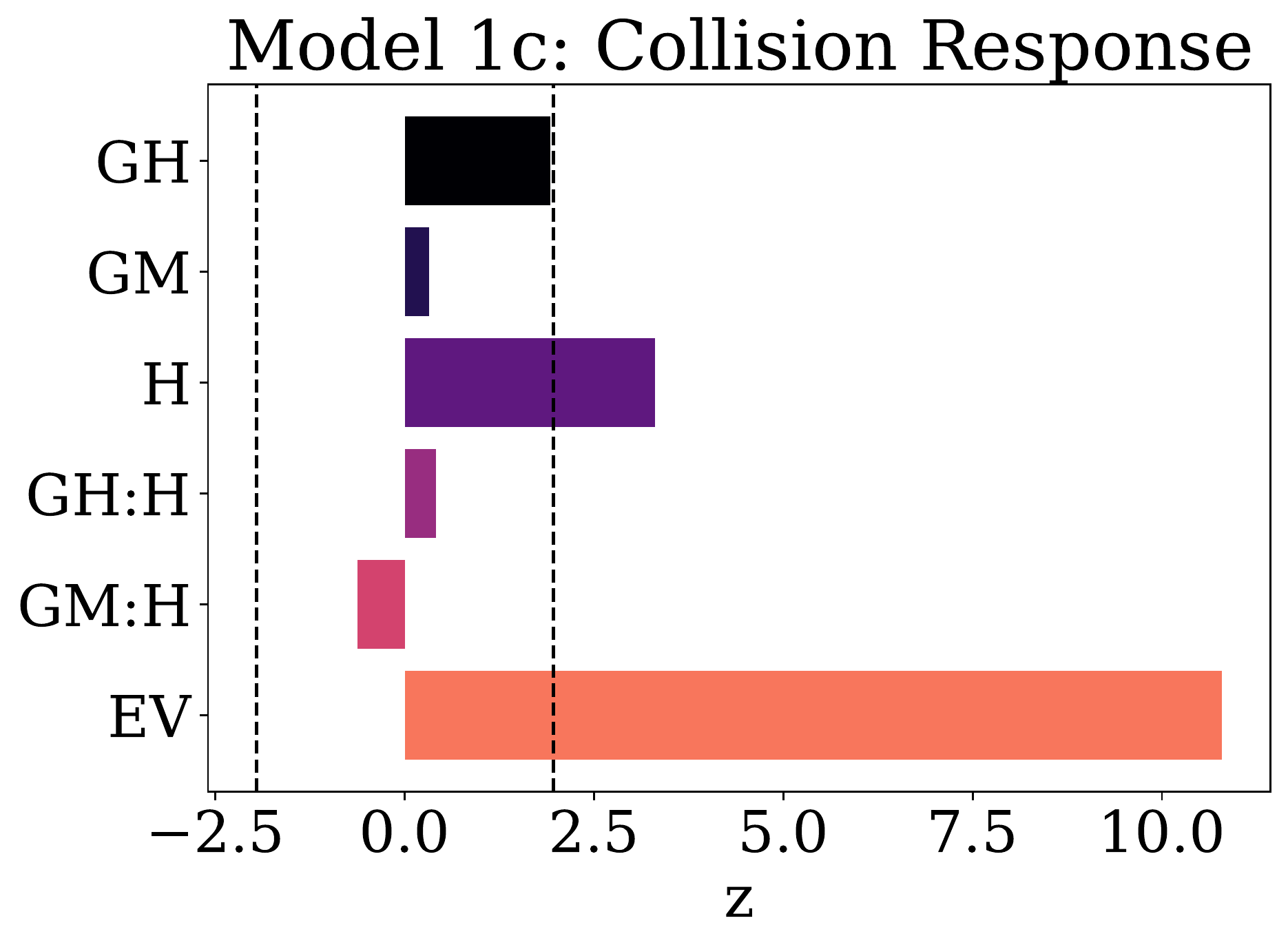}\par 
\end{multicols}
\begin{multicols}{2}
    \centering
     \includegraphics[width=\linewidth]{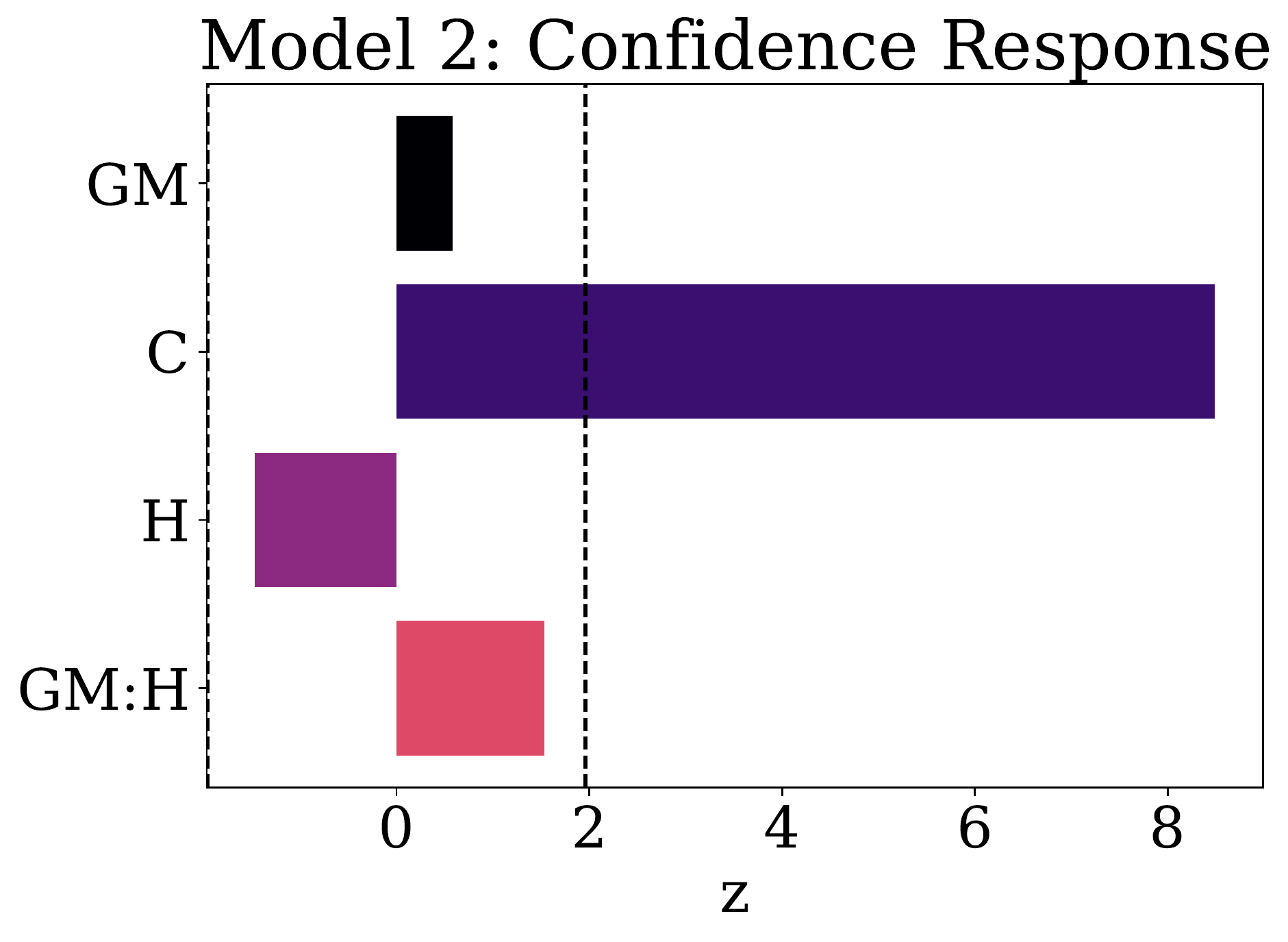}\par 
    \includegraphics[width=\linewidth]{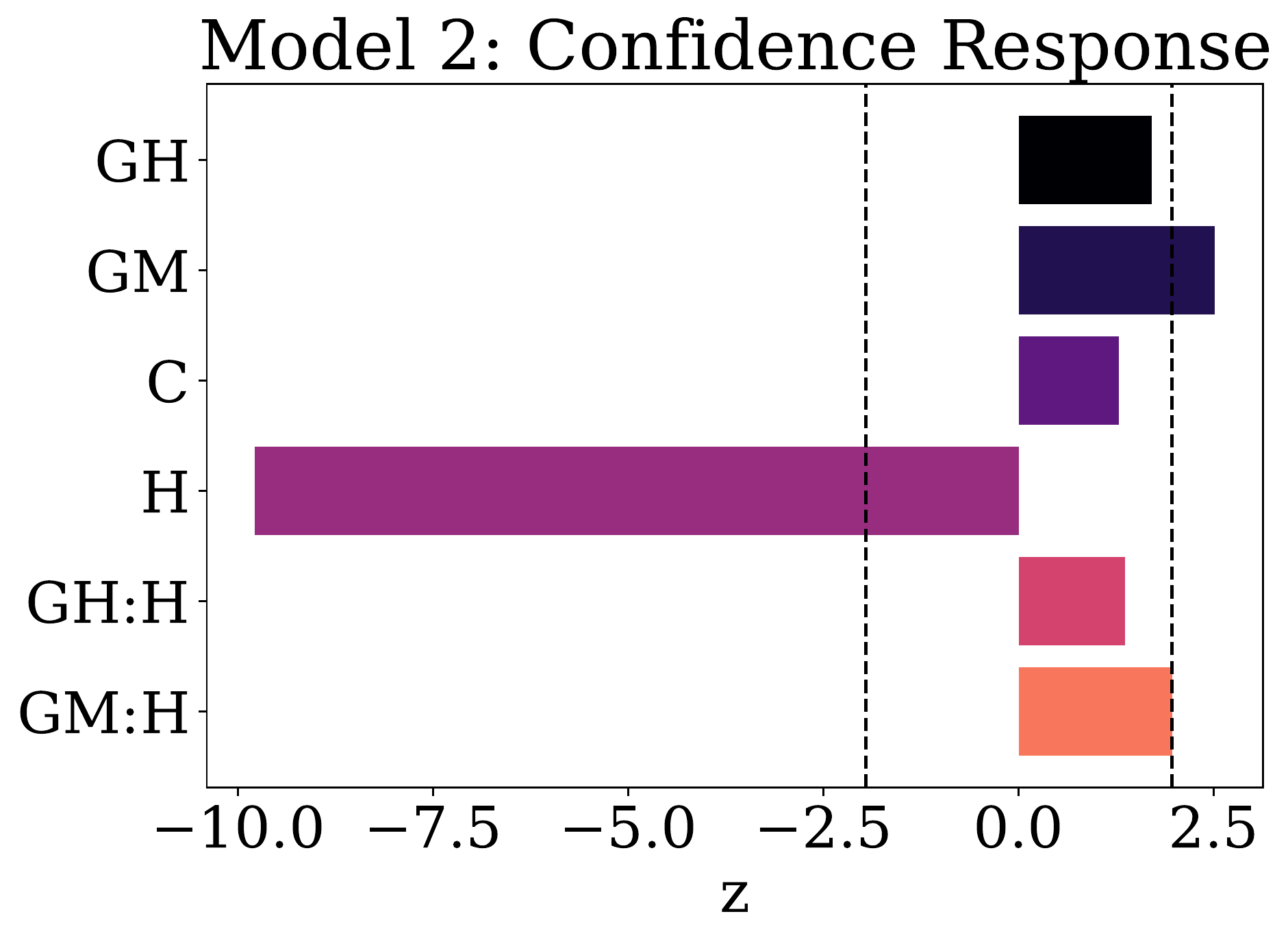}\par 
\end{multicols}
\caption{Tabular results of regression models in on-ground and parabolic flight experiments. Top row panels show coefficient values of effects on correct responses. Center row panels show coefficient values of effects on collision response choice. Bottom row panels show coefficient values of effects on confidence response. Critical values of two-tailed tests are represented by the vertical dashed lines. C: Correct response, EV: Expected value, GH: Hypergravity, GM: Microgravity, H: Uncertainty, RT: Response time, z: Z-statistic. }
\end{figure}

\section{Discussion}

This study investigated how gravity alterations affect decision making in human vision-based simulated drone navigation. Specifically, we queried how the latent variable of stimulus uncertainty in altered gravity affected different aspects of decision-making, including choice, accuracy and metacognitive confidence. We examined these questions first on the ground with simulated microgravity (Supine condition) using a visual self-motion estimation task in a VR environment before repeating the experiment in true micro- and hypergravity onboard a parabolic flight. The study questions were generated from a general, computational theory of brain function dubbed the predictive coding framework \cite{friston_does_2018}. Overall, we found no effect of gravity alterations on task performance relative to normal gravity. However, we found that microgravity boosted subjective confidence (metacognitive assessment of performance), especially under conditions of uncertainty. As the study was limited in sample size, these results remain preliminary and exploratory. However, if replicated, these findings can have 1) real world consequences on human spaceflight, and 2) highlight a potential gap to be solved with automated systems.

The on-ground experimental results provided support for our confirmatory hypotheses, namely that implicit uncertainty formalized as entropy prompts more errors in performance; demands a longer response time; and is reflected in subjective confidence reports. These results could fine-tune human-in-the-loop factors, for instance by designing systems that rely on intrinsic environmental uncertainty rather than a retrospective, human generated assessment. These results also provide a robust computational framework for further research investigating human performance in space analog environments. Uncertainty manipulation yielded expected effects, notably sigmoid curves linking responses between the two angle extremes (0 and maximum), as well as increased response ambivalence corresponding to the most ambiguous angles presented. In addition, this experiment showed a shift in confidence response curves relating to upward versus downward drone trajectory, indicative of an incongruence with upward crashes, and especially so in the Supine condition. The effect of supine position on visual perception of direction and self-motion estimation can introduce incongruence, as individuals in such cases rely on visual cues and acceleration, and less on vestibular information \cite{delle_monache_watching_2021}. In our study, external visual input was not available to subjects as they wore VR goggles throughout the experiment. Correct responses correlated with increased expected stimulus value, suggesting that higher stakes (probability of crash) may mobilize individuals towards accuracy, as seen in monetary decision-making \cite{shevlin_high-value_2022}. It is unclear as to why stimulus uncertainty led to no-crash responses. However, uncertainty in interaction with Supine condition did lead to an increase in crash predictions, underlining an effect of body position interacting with uncertainty to yield different choice behavior from the default upright.

No differences were found for gravity-related task performance on board the parabolic flight, supporting the overarching assumption of human adaptation to different environments \cite{tanaka_adaptation_2017}, \cite{carriot_challenges_2021}, \cite{scaleia_body_2019} though performance in-flight decreased considerably relative to the on-ground experiment. As in the on-ground experiment, stimulus uncertainty induced more mistakes but these did not vary as a function of gravity conditions. Stimulus uncertainty also induced longer reaction times, as expected \cite{bonnet_reaction_nodate}. As in the on-ground experiment, longer response times correlated with correct responses, indicative of an evidence accumulation process \cite{pereira_evidence_2021},  \cite{tagliabue_eeg_2019} and the expected value of the stimulus again prompted correct responses, suggesting an increase of effort or attention with higher-stakes \cite{reinhart_high_2014}. Stimulus uncertainty, in contrast to the on-ground experiment, predicted a crash response, in line with the notion that individuals are risk averse \cite{hintze_risk_2015},  \cite{kahneman2013prospect}. Interestingly, the hypergravity condition induced a marginal effect of crash prediction that, while not significant, may have come from the bodily sensation of the downward acceleration, usually predictive of an upcoming crash. Our confirmatory hypothesis with regards to subjective confidence reports were supported: stimulus uncertainty led to low confidence reports. However, microgravity conditions prompted high confidence reports, which may be linked an embodied feeling of expansion. The overconfidence effect was especially pronounced in microgravity with higher stimulus uncertainty, a marginal significant finding that may nonetheless hint at an alteration of uncertainty computation when the gravity prior is removed. 

Monitoring one’s performance can be described as assessing the degree to which one believes one is correct \cite{grimaldi_there_2015}. This ability is crucial for learning \cite{fromer_response-based_2021} and complex action such as sports performance \cite{locke_performance_2020}. Little research has been conducted on how metacognition varies in different environments. One study found that time spent in extreme altitudes (Mt. Everest specifically) caused no impairment in performance or response time but did impact metacognitive assessments \cite{nelson_cognition_1990}. Human metacognitive lability may be rescued by autonomous agents or perhaps with training \cite{frye_what_2014}, two solutions that may complement each other, as a uniquely autonomous solution may prompt unintended consequences stemming from a decrement in human learning \cite{kayes_cognitive_2022}. Several papers have identified broad categories of human factors to consider in extreme environments such as mountaineering \cite{wickens_human_2015}, military operations \cite{militello_designing_2015}, and spaceflight \cite{orasanu_crew_2005}. In our study, microgravity led to increases in confidence reports, a deviation that may come directly from gravity changes or through an indirect, embedded mechanism, such as affect or body sensation. Subjects reported euphoria in weightlessness, a known effect from parabolic flight experiments \cite{gerathewohl_personal_1956} and subjective positive affect is known to induce increases in confidence \cite{molenberghs_neural_2016}. While individuals adapt to differences in vestibular information \cite{carriot_challenges_2021} and gravity differences \cite{goswami_human_2021}, \cite{tanaka_adaptation_2017}, it is also widely reported that astronauts experience a lasting “space euphoria” \cite{hupfeld_microgravity_2021}. If euphoria both leads to overconfidence and simultaneously does not dampen over time, it may effect decision-making and error monitoring. A sense of body expansion may have also contributed to alterations in judgment \cite{tajadura2018audio}. In addition and more importantly, based on our results, microgravity, or its affective  or interoceptive features, may interfere with uncertainty processing specifically, leading to a risk discounting scenario that may, in situations of danger, be maladaptive. 

The study was subject to several limitations common to space analog research, chief among them a low sample size of 2 but also, due to the limited number of parabolas experienced (20 across the two participants), as well as the limited duration of the altered gravity epoch, a low number of trials experienced in altered gravity. Acquiring empirical data on human performance in extreme environments is a uniquely challenging endeavor and sample sizes therefore remain limited \cite{ploutz2014justifying}. Replications both in future studies and also across different perceptual and decision-making tasks can overcome this shortcoming. In addition, the role of mood and affect were not foreseen to play a role in study outcomes, and therefore were not explicitly assessed. These unique human factors may play an important role in the results observed and should be formally measured in future studies.

Decision-making under uncertainty is exceptionally important to manned space missions. Given the considerable cost of choosing the wrong option, it is incumbent on researchers to consider myriad factors influencing decision-making in space including isolation, confinement, sleep deprivation, stress, and gravity changes. The effects of all but the last can easily be simulated on Earth. Exploring human behavioral responses to uncertainty in parabolic flight informs on gravity's unique contribution to human decision-making capabilities. Future studies can further explore associated dimensions to decision-making, including affect, interoception and attention \cite{pagnini2023placebo}. While parabolic flights present constraints with respect to time and study sample, they provide a snapshot into how best to design Mars-bound drones, support human operators, and more generally provide insight on how humans resolve gravity changes in dynamic environments, with a view towards addressing some questions relating to (hu)manned missions to space.

\section{Methods}

\subsection{Ethics statement}

The experimental protocol was approved by the local ethics committee of the Department of Economics at the University of Zurich. 
All experiments were performed in accordance with relevant guidelines and regulations.
Written informed consent was obtained from all participants before participating in the study.
Each participant received a full debrief and a monetary compensation for their time of 25 CHF/hour after the experiment.

\subsection{Participants}

A total of 22 volunteers (6 female, median age: 25.6 years, age range: 22 - 43 years) were recruited for the study. 
Handedness was assessed with the Edinburgh Handedness Inventory (Oldfield 1971). Out of 22 subjects, 16 were right-handed (Laterality Quotient between 61 and 100), 6 were ambidextrous (Laterality Quotient between -60 and 60), and 0 were left-handed (Laterality Quotient between -100 and -61). 
All participants were healthy without prior history of psychological or neurological impairments, according to self-report. 
The study was conducted at the Robotics and Perception Group laboratory at the University of Zurich. 
Participants were asked to read and sign information and consent forms prior to the experiment. 
Participants were then asked to provide biographical information, including handedness and video-game-playing experience.

\subsection{Self motion estimation experiment: procedure, stimuli and setup}

Two experiments were conducted to address our research questions, one on-ground and the other in-flight. In both studies, participants performed a self-motion estimation task, presented through virtual reality (VR) goggles (DJI FPV Goggles, \href{https://dji.com/}{https://dji.com/}). An IMU recorder was also affixed to the VR goggles to log exact acceleration forces. Within the VR environment, participants were tasked with making a perceptual decision when viewing a drone moving through and out of a cave from a first person perspective. Specifically, participants were asked to predict a collision (or no collision) with either the ceiling or floor of the cave after 1.5 s of motion and just prior to the drone’s egress from the cave. Following their collision response, participants had to provide a confidence estimate on their response on a scale ranging from 0 to 10. 
The angle of the drone’s trajectory was pseudo-randomly chosen at each trial.
Drone trajectory angles ranged from 19.58\textdegree downward pitch to 19.58\textdegree  upward pitch, with a pitch angle difference of 1.96\textdegree  between trajectories. 
There were thus 21 trajectories that were rendered in Unity software (\href{https://unity.com/}{https://unity.com/}, embedded in PsychoPy (\href{https://www.psychopy.org/}{https://www.psychopy.org/} and presented on a Lenovo W541 laptop. Responses were recorded via a Sony Playstation game controller.  A total of 420 trials were performed, chunked in 4 blocks. In between blocks, participants were asked to complete a brief questionnaire (NASA TLX, \cite{hart_tlx_1986, hart_tlx_2006}) to index their subjective assessment of task difficulty. 

\begin{figure*}[!ht]
\centering
\begin{center}
\includegraphics[trim = 0mm 0mm 0mm 0mm, clip, width=\linewidth]{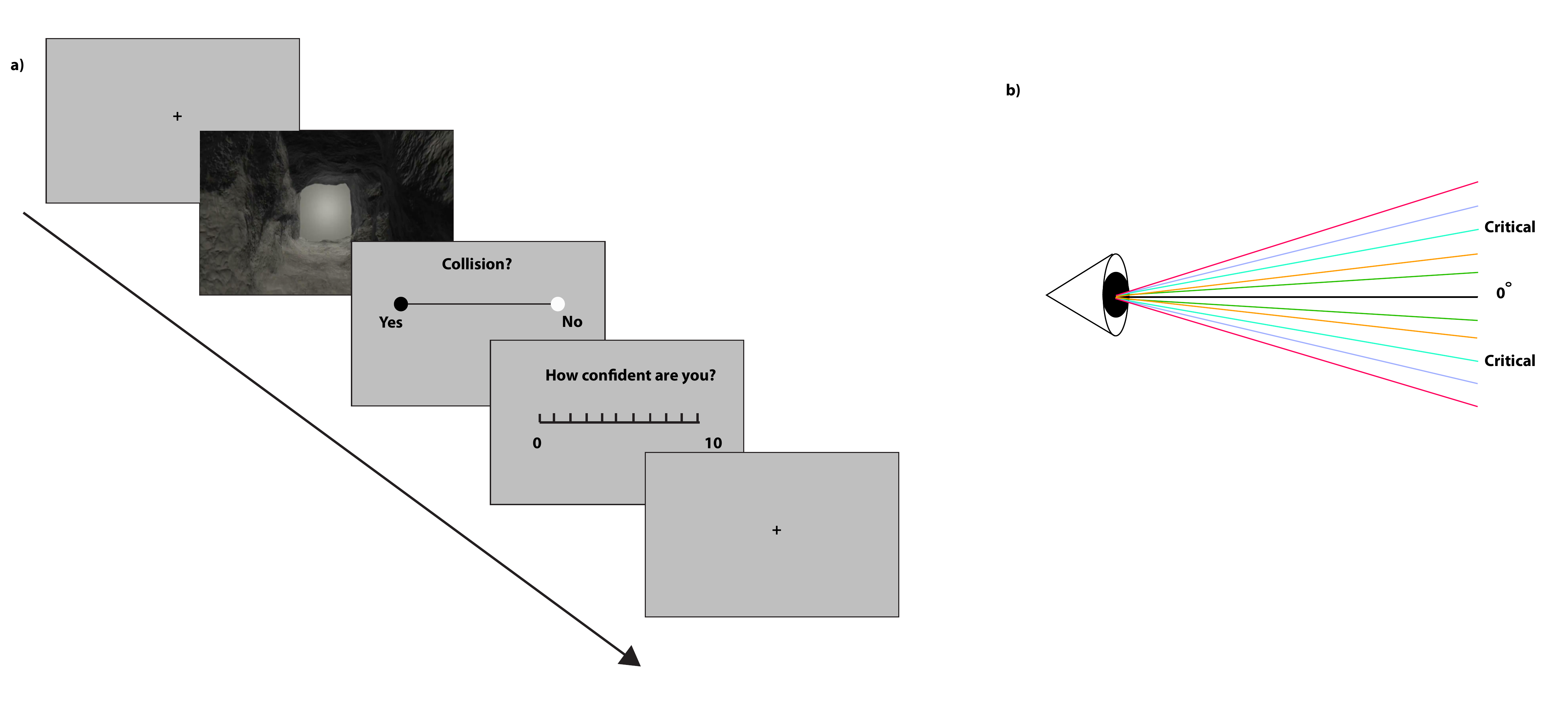}  
\caption[caption]{Schematic representation of a sample trial and stimuli. a) Trial progression. Participants first view 1.5 s of a video from a first person perspective, where they are moving towards a cave exit. They are then asked to predict whether they would have crashed or not. In the next step, they are asked to provide an estimate of their confidence in their response. b) Representation of uncertainty induction. Videos were angled at successive intervals from a 0\textdegree angle point-of-view (head-on vision). Critical values are angles that, in either up or down directions, lie halfway between a sure, crash-free exit (0\textdegree) and a sure crash (highest absolute angle deviating from 0)}
\vskip -0.3in 
\label{fig:parabolic-psychometric-fitting-results}
\end{center}
\end{figure*}

\subsection{Parabolic flight campaign}

Two participants from the on-ground experiment cohort (1 male, 1 female) went on to perform a similar experiment on board a Cessna Citation-II (Figure 1) that underwent 20 parabolic maneuvers over 2 days. In-flight, the experiment was designed to be continuous and included a limited range of collision trajectory angles that corresponded to subjective probabilities of collision extracted from the OG study curves for the two participants. Stimuli corresponding to a crash probability \emph{p} of 0, 0.25, 0.5, 0.75 and 1  were employed in the parabolic flight experiment. As in the on-ground experiment, participants wore the same VR goggles with the IMU sensor attached and logging responses with the same controller. In addition, subjects wore seatbelts during the experiment, limiting their movement. The experiment started once the aircraft reached cruising altitude. On day 1, flight 1, Subject 1 performed the experiment across 5 parabolas before the flight was cut short and forced to land. On day 2, flight 2, subject 1 performed the experiment across an additional 5 parabolas, and subject 2, across 10 parabolas. On the climb and descent from the apex of the parabola, gravity forces reached 2.5 G; at the summit, 0 G. Hypergravity phases lasted approximately 20 s and microgravity, 30 s.

\subsection{Analysis}

\subsubsection{Data pre-processing} 

Of 22 participants, one was excluded due to a technical problem with the response box, where ratings were not recorded.
The remaining data from 21 participants were used for subsequent analysis.
Raw data inspection showed that most response times (RTs) distributions follow an expected gamma distribution. We therefore compute inverse RTs (formula: 1/RT) to convert the distribution to a gaussian before their inclusion in analysis models. We removed trials for which RTs or inverse RTs exceeded 5 times the interquartile range around the individual subject’s median. A maximum of 16\% of trials per subject were removed, leading to 230-251 accepted trials per subject, resulting in a total of 4828 accepted trials for 21 participants used for subsequent analysis. Data streams from the betaflight IMU were time synchronized to the experiment computer clock by correlating the norm accelerometer data, determining the time-offset at peak correlation, and correcting for the time offset. In the parabolic flight experiment, trials were labeled normogravity when the median of the norm accelerometer recording during the video presentation period was between 0.5 - 1.5 G. Trials were labeled hypergravity if this metric exceeded 1.5 G, and microgravity at less than 0.5 G.

\subsubsection{Data analysis}

Summary statistics collected included crash prediction response frequency; performance accuracy across conditions; and average confidence reports across conditions, in both on-ground and parabolic flight experiments.

A main goal of the study was to determine if gravity interacts with latent variables relating to decision-making under uncertainty to affect perceptual decisions within a predictive coding framework. We first compute key variables that capture the probabilistic value and associated uncertainty of each video. Mathematically, this yields the expected value of a given stimulus (collision weighted by the subjective probability of collision for that video, Eq. 1); and the entropy of that same stimulus (Eq. 2). None of these values are thought to be computed in an explicit manner, which emphasizes their characterization as hidden variables.

\subsubsection{Predictive coding model: expected stimulus value and stimulus uncertainty}

A full predictive coding model consists of two epochs in a given process: 1) a prediction and its associated variables; 2) an update and its associated variables. We focus on the first epoch in our experiment, where we consider the expected value of the stimulus as well as the prediction's uncertainty. 

As each drone angle was presented 10 times to each subject, a probability of responding one way or another was computed for each unique stimulus by dividing the sum of crash predictions \emph{c} by \emph{K} the number of times the stimulus was presented (10).

\begin{equation}
p = \frac{\sum c}{K} \times{1}
\end{equation}

Stimulus uncertainty in the form of entropy was thereafter derived for each stimulus as: 

\begin{equation}
H = -p \times \log_{2} p - q \times \log_{2} q
\end{equation}

Where \emph{p} is the probability of a collision and \emph{q}, the probability of no collision. All \emph{p} and \emph{q} $\notin(0,1)$, were replaced by 0.001 and 0.999, respectively.

\subsubsection{General linear models}

We perform 4 general linear mixed models on the data collected to detect gravity effects on correct response; collision prediction; collision prediction reaction time; and confidence rating; with subject identity included as a random effect. In the parabolic flight experiment, because of the low n (2), we do not employ a mixed model with subject as a random factor in the regression analyses.

\subsection{Data availability statement}

The datasets generated during the current study are available in from the following OSF repository: \protect\url{https://osf.io/zqfw4/}.

\subsection{Competing interests statement}

The authors declare no competing interests.

\section{Author contributions}
L.L.K. and C.P.(study conception, study design, computational and statistical analyses, data interpretation, writing); R.S. and S.A. (data collection and analysis); D.S. (writing and funding)

{\small
\bibliographystyle{IEEEtran}
\bibliography{references}
}

\end{document}